\documentclass[journal, twoside, twocolumn]{IEEEtran}

\ifCLASSOPTIONcompsoc
  \usepackage[nocompress]{cite}
\else
  \usepackage{cite}
\fi

\ifCLASSINFOpdf
 \else
 \fi

\usepackage{algorithm}
\usepackage{algorithmic}

\usepackage{soul}
\usepackage{mathrsfs}
\usepackage{amssymb}
\usepackage{amsmath}
\usepackage{dsfont}
\usepackage{booktabs}
\usepackage{epstopdf}
\usepackage{multirow}
\usepackage{comment}
\usepackage{endnotes}
\usepackage{hyperref}
\usepackage{bibentry}
\usepackage{amsmath}
\usepackage{dsfont}
\usepackage{amsfonts}
\usepackage{amssymb}
\usepackage{bbm}
\usepackage{bbding}
\usepackage{makecell}
\usepackage[normalem]{ulem}

\usepackage{amsmath}
\usepackage{tikz}
\usepackage{array}
\usepackage{times}
\usepackage{subfig}
\usepackage{helvet}
\usepackage{subfig}
\usepackage{amssymb}
\usepackage{amsmath}
\usepackage{amssymb}
\usepackage{courier}
\usepackage{multirow}
\usepackage{mathrsfs}
\usepackage{graphicx}
\usepackage{enumitem}
\usepackage{graphicx}
\usepackage{blindtext}
\usepackage{algorithm}
\usepackage{algorithmic}
\usepackage{lineno,hyperref}
\usepackage[marginal]{footmisc}
\usepackage[utf8]{inputenc}
\usepackage[english]{babel}
\usepackage{epstopdf}
\usepackage{array}
\usepackage{multirow}
\usepackage{booktabs}

\begin{document}
\title{Simple Contrastive Graph Clustering}

\author{Yue~Liu,~Xihong~Yang,~Sihang~Zhou,~Xinwang~Liu
\IEEEcompsocitemizethanks{
\IEEEcompsocthanksitem Y. Liu, X. Yang, X. Liu, S. Wang are with College of Computer Science and Technology, National University of Defense Technology, Changsha, 410073, China. (E-mail: yueliu@nudt.edu.cn)


}
}

\markboth{UNDER REVIEW}%
{Y. Liu \MakeLowercase{\textit{et al.}}: Simple Contrastive Graph Clustering}

\maketitle

\begin{abstract}
Contrastive learning has recently attracted plenty of attention in deep graph clustering for its promising performance. However, complicated data augmentations and time-consuming graph convolutional operation undermine the efficiency of these methods. To solve this problem, we propose a Simple Contrastive Graph Clustering (SCGC) algorithm to improve the existing methods from the perspectives of network architecture, data augmentation, and objective function. As to the architecture, our network includes two main parts, i.e., pre-processing and network backbone. A simple low-pass denoising operation conducts neighbor information aggregation as an independent pre-processing, and only two multilayer perceptrons (MLPs) are included as the backbone. For data augmentation, instead of introducing complex operations over graphs, we construct two augmented views of the same vertex by designing parameter un-shared siamese encoders and perturbing the node embeddings directly. Finally, as to the objective function, to further improve the clustering performance, a novel cross-view structural consistency objective function is designed to enhance the discriminative capability of the learned network. Extensive experimental results on seven benchmark datasets validate our proposed algorithm's effectiveness and superiority. Significantly, our algorithm outperforms the recent contrastive deep clustering competitors with at least seven times speedup on average.
\end{abstract}


\section{INTRODUCTION}
Graph learning \cite{tang_1,tang_2,pan_1,CG3} is becoming increasingly crucial in many applications like facial expression recognition \cite{face_1}, video action recognition \cite{video_1}, and the recommendation system \cite{recommendation_1} for its good hidden correlation exploiting capability. Among all the directions in graph learning, a fundamental and challenging task, i.e., deep graph clustering, has recently attracted intensive attention \cite{DAEGC,ARGA,ARGA_conf,SDCN,DFCN,AGCN,AGE,MVGRL,MCGC,DCRN,SCAGC,MGCCN,DNENC,FGC}.



%


According to the learning mechanism, the existing deep graph clustering methods can be roughly categorized into three classes: generative methods \cite{GAE,MGAE,DAEGC,AGC,GALA,MAGCN,AGCN,SDCN,DFCN}, adversarial methods \cite{ARGA,ARGA_conf,AGAE}, and contrastive methods \cite{MVGRL,AGE,DCRN,MCGC,MGCCN,SCAGC,GDCL}. In early literature, the generative methods and adversarial methods improve clustering performance by learning cluster-oriented node representations and designing fake sample generation-recognition mechanisms, respectively. However, since most of these methods adopt a clustering guided loss function \cite{DEC} to force the generated sample embeddings to have the minimum distortion against the pre-learned clustering centers \cite{DAEGC,MAGCN,AGCN,ARGA,ARGA_conf,AGAE,SDCN,DFCN}, their clustering performance is highly dependent on good initial cluster centers, thus leading to manual trial-and-error pre-training. As a consequence, the performance consistency, as well as the implementing convenience, is largely decreased. More recently, thanks to the development of contrastive learning, more consistent and discriminative contrastive loss functions are designed to replace the clustering guided loss function for network training. As a result, the manual trial-and-error problem is alleviated, and the clustering performance is improved \cite{MVGRL,AGE,DCRN,MCGC,MGCCN,SCAGC,GDCL}. However, complicated data augmentations and time-consuming graph convolutional operation undermine the efficiency of these methods, making them computational time and space consuming (See Section \ref{costs_sec}).

To solve these problems, we propose a Simple Contrastive Graph Clustering (SCGC) method to improve the existing methods from the aspect of network architecture, data augmentation, and objective function. To our network architecture, the backbone is designed with a siamese network whose sub-branch merely consists of the MLP. The neighborhood information aggregation process is conducted independently before network training. In this manner, we filter the high-frequency noise in attributes, thus improving both the clustering performance and training efficiency. (See Section \ref{low_pass_SCM_sec} and Section \ref{costs_sec}). For data augmentation, instead of constructing two different views of the same node with complex modification against graphs, we implement it by designing parameter un-shared siamese encoders and perturbing the node embeddings with Gaussian noise directly. Moreover, we design a neighbor-oriented contrastive objective function to force the cross-view similarity matrix to approximate the self-looped adjacency matrix. By this setting, the network is endowed with the ability to keep the cross-view structural consistency, thus further improving the clustering performance (See Section \ref{low_pass_SCM_sec}). Benefiting from our simple architecture, SCGC is free from pre-training and outperforms the recent contrastive competitors with at least seven times speedup on average. Meanwhile, we save about 59\% GPU memory against other contrastive methods on average (See Section \ref{costs_sec}). The main contributions of this paper are summarized as follows.

\begin{table*}[]
\caption{Differences between our proposed SCGC and other contrastive deep graph clustering methods from perspectives of data augmentation, network architecture, and objective function.}
\small
\scalebox{1.07}{
\begin{tabular}{cccc}
\hline
  Method    & Data Augmentation                    & Network Architecture & Objective Function Form                   \\ \hline
AGE \cite{AGE}   & No Data Augmentation                                    & Graph Filters+MLP  & Cross-Entropy Loss               \\
MVGRL \cite{MVGRL} & Graph Diffusion                      & Un-shared GCNs+ shared MLP           & InfoMax Loss \\

MCGC \cite{MCGC} & No Data Augmentation                                    & Graph Filters      & InfoNCE Loss                     \\
DCRN \cite{DCRN} & Attribute Perturbation+Graph Diffusion        & Auto-Encoder+GCN  & MSE Loss to Identity Matrix          \\
SCAGC \cite{SCAGC} & Attribute Perturbation+Edge Perturbation & Shared GCN               & InfoNCE Loss                     \\
MGCCN \cite{MGCCN} & No Data Augmentation                                    & Shared GCN               & InfoNCE Loss                     \\
SCGC (Ours)   & Un-shared MLPs+Embedding Perturbation         & Graph Filters+Un-shared MLPs  & MSE Loss to Adjacent Matrix           \\ \hline
\end{tabular}}
\label{different}
\end{table*}

\begin{itemize}
\item We propose a simple yet effective contrastive deep graph clustering method termed SCGC. Benefit from its simplicity, SCGC is free from pre-training and saves both time and space for network training.

\item A new data augmentation method, which conducts data perturbation only in the enhanced attribute space, is proposed. This fashion is verified to be compatible with the existing contrastive methods.



\item We design a novel neighbor-oriented contrastive loss to keep the structural consistency even across views, thus improving the discriminative capability of our network.

\item Extensive experimental results on seven benchmark datasets demonstrate the superiority and efficiency of the proposed method against the existing state-of-the-art deep graph clustering competitors.

\end{itemize}

\section{RELATED WORK}

\subsection{Deep Graph Clustering}
Deep graph clustering is a fundamental yet challenging task that aims to reveal the underlying graph structure and divides the nodes into several disjoint groups. According to the learning mechanism, the existing deep graph clustering methods can be roughly categorized into three classes: generative methods \cite{GAE,MGAE,DAEGC,AGC,GALA,MAGCN,AGCN,SDCN,DFCN}, adversarial methods \cite{ARGA,ARGA_conf,AGAE}, and contrastive methods \cite{MVGRL,AGE,DCRN,MCGC,MGCCN,SCAGC,GDCL}. Our proposed method belongs to the last category. We will review the generative methods and adversarial methods in this section and detail the difference between our proposed method and other contrastive methods in the next section.

The pioneer graph clustering algorithm MGAE \cite{MGAE} embeds nodes into the latent space with GAE \cite{GAE} and then performs clustering over the learned node embeddings. Subsequently, DAEGC \cite{DAEGC} and MAGCN \cite{MAGCN} improve the clustering performance of early works with the attention mechanisms \cite{attention,GAT}. Besides, GALA \cite{GALA} and AGC \cite{AGC} enhance the GAE by the symmetric decoder and the high-order graph convolution operation, respectively. In addition, ARGA \cite{ARGA,ARGA_conf} and AGAE \cite{AGAE} improve the discriminative capability of samples through adversarial mechanisms \cite{GAN,graph_GAN}. Moreover, SDCN \cite{SDCN}, AGCN \cite{AGCN}, and DFCN \cite{DFCN} verify the effectiveness of the attribute-structure fusion mechanisms to improve the clustering performance.

Although verified to be effective, since most of these methods adopt a clustering guided loss function \cite{DEC} to force the learned node embeddings to have the minimum distortion against the pre-learned clustering centers, their clustering performance is highly dependent on good initial cluster centers, thus leading to manual trial-and-error pre-training \cite{DAEGC,MAGCN,AGCN,ARGA,ARGA_conf,AGAE,SDCN,DFCN}. As a consequence, the performance consistency, as well as the implementing convenience, is largely decreased. Unlike them, our proposed method replaces the clustering guided loss function by designing a novel neighbor-oriented contrastive loss function, thus getting rid of trial-and-error pre-training.

\subsection{Contrastive Deep Graph Clustering}
Contrastive learning has achieved great success on images \cite{DIM,SIMCLR,BYOL,BARLOW,GCC,xihong} and graphs \cite{DGI,GraphCL,GRACE,simgrace,GRAPH_BYOL,GBT,graph_mlp} in recent years. Inspired by their success, contrastive deep graph clustering methods \cite{AGE,MVGRL,MCGC,SCAGC,DCRN,IDCRN,GDCL,MGCCN} are increasingly proposed.

Three key factors, i.e., data augmentation, network architecture, and objective function, significantly determine the clustering performance of the contrastive methods. According to these factors, we summarize the differences between our proposed SCGC and other contrastive deep graph clustering methods in Table \ref{different}.

\noindent{\textbf{Data augmentation.}} The existing data augmentations in contrastive methods aim to build different views of the same vertex by introducing complex operations over graphs. Specifically, MVGRL \cite{MVGRL} and DCRN \cite{DCRN} adopt the graph diffusion matrix as an augmented graph. Besides, SCAGC \cite{SCAGC} perturbs the graph topology by randomly adding or dropping edges. In addition, DCRN and SCAGC conduct augmentations on node attributes by the attribute perturbation. Although verified to be effective, these data augmentations are complicated and still entangle the aggregation and transformation during training, thus limiting the efficiency of the contrastive methods. Different from them \cite{MVGRL,SCAGC,DCRN}, our SCGC constructs two augmented views of the same vertex by simply designing parameter un-shared siamese encoders and perturbing embeddings directly instead of introducing any complex operations over graphs.


\noindent{\textbf{Network architecture.}} To the network architecture, SCAGC \cite{SCAGC} and MGCCN \cite{MGCCN} both encode nodes with the shared GCN encoders \cite{GCN}. Differently, MVGRL \cite{MVGRL} adopt two-parameter un-shared GCN encoders and the shared MLP as the backbone. In addition, DCRN \cite{DCRN} utilizes auto-encoder \cite{AE_K_MEANS} and GCN encoder to embed augmented views into the latent space. However, previous GCN encoders all entangle the transformation and aggregations operation during training, thus leading to high time costs. To solve this issue, AGE \cite{AGE} decouple these two operations in GCN by a graph Laplacian filter \cite{lapalician-filter} and one MLP. Different from AGE \cite{AGE}, we encode the smoothed node attributes with two separated MLPs, which have the same architecture but un-shared parameters.

\noindent{\textbf{Objective function.}} Thirdly, for the objective function, MVGRL \cite{MVGRL} designs the InfoMax loss \cite{DIM} to maximize the cross-view mutual information between the node and the global summary of the graph. Meanwhile, AGE \cite{AGE} designs a pretext task to classify the similar nodes and the dissimilar nodes by the cross-entropy loss. Subsequently, SCAGC \cite{SCAGC}, MCGC \cite{MCGC}, and MGCCN \cite{MGCCN} all adopt the infoNCE loss \cite{infoNCE} to pull together the positive sample pairs while pushing away the negative sample pairs. Concretely, based on similarity, MCGC defines the positive samples as the k-nearest neighbors of the node while regarding other nodes as negative samples. SCAGC designs the contrastive clustering loss to maximize the agreement between representations of the same cluster. MGCCN pulls close the embeddings of the same node in different GCN layers and pushes away the embeddings of different nodes. In addition, DCRN designs the MSE loss to reduce the redundancy in the feature level and sample level. Different from them, we design a novel neighbor-oriented contrastive loss to keep the structural consistency even across views, thus improving the discriminative capability of our network.

\begin{figure*}
\centering
\includegraphics[scale=0.72]{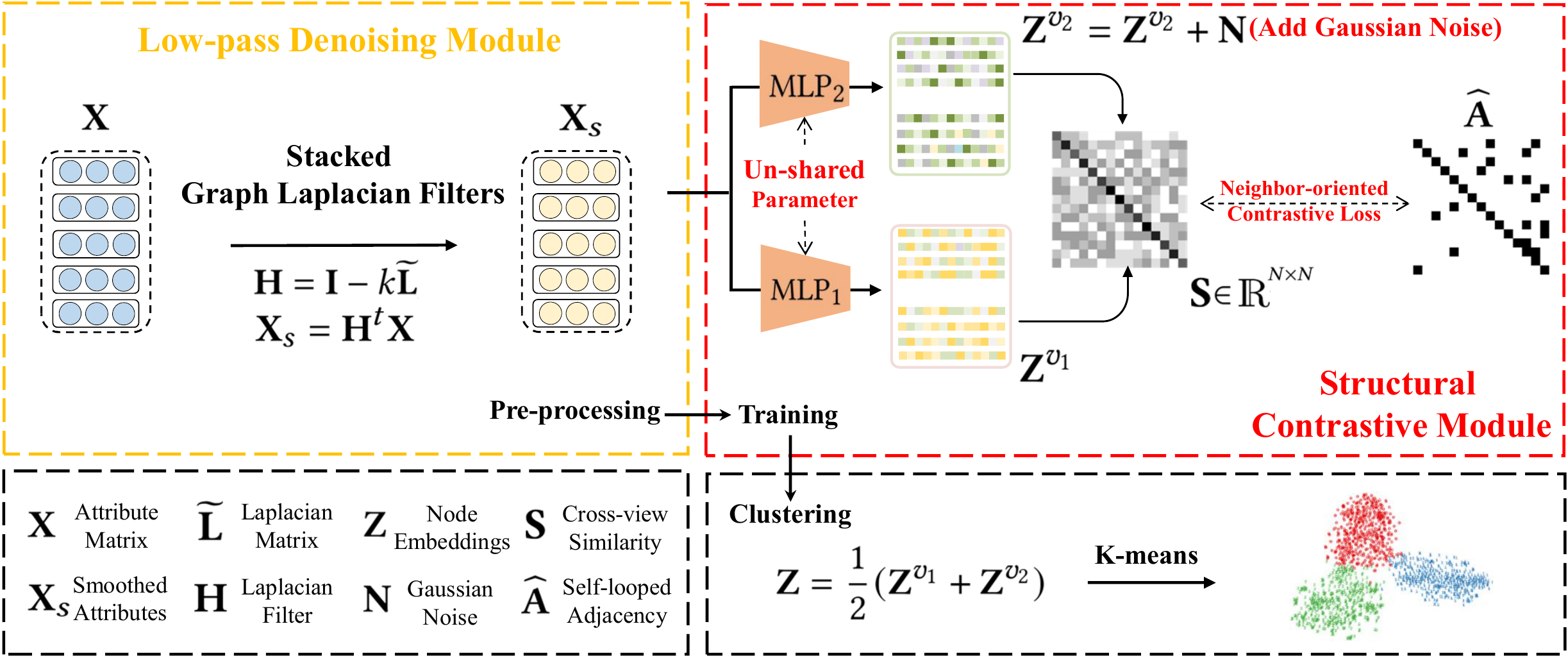}
\caption{Illustration of the Simple Contrastive Graph Clustering (SCGC) algorithm. In our proposed algorithm, we firstly pre-process the node attributes by the low-pass denoising operation. Then, the structural contrastive module encodes the smoothed node attributes with merely two MLPs, and constructs augmented views of node embeddings by designing parameter un-shared siamese encoders and perturbing the node embeddings. Moreover, a novel neighbor-oriented contrastive loss is designed to keep the cross-view structural consistency, thus improving the discriminative capability of the network.}
\label{OVERRALL_FIGURE}  
\end{figure*}





 

 


\begin{table}[h]
\centering
\caption{Notation summary.}
\small
\scalebox{1.1}{
\begin{tabular}{ll}
\toprule
\textbf{Notation}                                        & \textbf{Meaning}                                \\ \midrule
$\textbf{X}\in \mathds{R}^{N\times D}$  & Attribute matrix  
\\
$\textbf{X}_s\in \mathds{R}^{N\times D}$  & Smoothed attribute matrix  
\\
$\textbf{A}\in \mathds{R}^{N\times N}$  & Original adjacency matrix   
\\
$\widehat{\textbf{A}}\in \mathds{R}^{N\times N}$  & Adjacency matrix with self-loop
\\
$\textbf{I} \in \mathds{R}^{N\times N}$  & Identity matrix  
\\     
$\textbf{D}\in \mathds{R}^{N\times N}$  & Degree matrix         
\\
$\textbf{L}\in \mathds{R}^{N\times N}$  & Graph Laplacian matrix
\\
$\widetilde{\textbf{L}}\in \mathds{R}^{N\times N}$  & Symmetric normalized Laplacian matrix
\\
$\textbf{Z}^{v_k} \in \mathds{R}^{N\times d}$   & Node embeddings in $k$-th view      
\\
$\textbf{S} \in \mathds{R}^{N\times N}$  & Cross-view sample correlation matrix  
\\     
$\textbf{Z} \in \mathds{R}^{N\times d}$  & Clustering-oriented node embeddings      
\\
\bottomrule
\end{tabular}
}
\label{NOTATION_TABLE} 
\end{table}

\section{METHODOLOGY}
\subsection{Notations and Problem Definition}
Let $\mathcal{V}=\{v_1, v_2, \dots, v_N\}$ be a set of $N$ nodes with $C$ classes and $\mathcal{E}$ be a set of edges. In the matrix form, $\textbf{X} \in \mathds{R}^{N\times D}$ and $\textbf{A} \in \mathds{R}^{N\times N}$ denote the attribute matrix and the original adjacency matrix, respectively. Then $\mathcal{G}=\left \{\textbf{X}, \textbf{A} \right \}$ denotes an undirected graph. The degree matrix is formulated as $\textbf{D}=diag(d_1, d_2, \dots ,d_N)\in \mathds{R}^{N\times N}$ and $d_i=\sum_{(v_i,v_j)\in \mathcal{E}}a_{ij}$. The graph Laplacian matrix is defined as $\textbf{L}=\textbf{D}-\textbf{A}$. With the renormalization trick $\widehat{\textbf{A}}=\textbf{A}+\textbf{I}$ in GCN \cite{GCN}, the symmetric normalized graph Laplacian matrix is denoted as $\widetilde{\textbf{L}} = \widehat{\textbf{D}}^{-\frac{1}{2}}\widehat{\textbf{L}}\widehat{\textbf{D}}^{-\frac{1}{2}}$. The notations are summarized in Table \ref{NOTATION_TABLE}.

Deep graph clustering aims to divide the nodes in the graph into several disjoint groups in an unsupervised manner. Concretely, a neural network $\mathcal{F}$ is firstly trained in an unsupervised manner and encodes the nodes by exploiting node attributes and structural information as follows:

\begin{equation} 
\textbf{E} = \mathcal{F}(\textbf{A}, \textbf{X}),
\label{Embed}
\end{equation}
where \textbf{X} and \textbf{A} denotes the attribute matrix and the original adjacency matrix. Besides, $\textbf{E} \in \mathds{R}^{N \times d}$ is the learned node embeddings, where $N$ is the number of samples and $d$ is the number of feature dimensions. After that, a clustering algorithm $\mathcal{C}$ such as K-means\cite{KMEANS}, 
spectral clustering \cite{spectral_clustering}, or clustering neural network \cite{SDCN} is adopted to devides learned node embeddings $\textbf{E}$ into $k$ disjoint groups as follows:

\begin{equation} 
\Phi = \mathcal{C}(\textbf{E}),
\label{clustering}
\end{equation}
where $\Phi \in \mathds{R}^{N \times k}$ denotes the cluster membership matrix for all $N$ nodes.

\subsection{Overall Framework}
We propose a Simple Contrastive Graph Clustering (SCGC) algorithm. The framework of SCGC is shown in Fig. \ref{OVERRALL_FIGURE}. It mainly consists of two components: low-pass denoising operation and Structural Contrastive Module (SCM). In the following sections, we will detail low-pass denoising operation, SCM, and the objective function.


\subsection{Low-pass Denoising Operation}
Recent works \cite{AGE,SGC,Deeper_insights} have demonstrated that the Laplacian filter \cite{lapalician-filter} can achieve the same effect as the graph convolution operation \cite{GCN}. Motivated by their success, we introduce a low-pass denoising operation to conduct neighbor information aggregation as an independent pre-processing before training. In this manner, the high-frequency noise in attributes will be filtered out efficiently.


Concretely, we introduce a graph Laplacian filter as formulated:
\begin{equation} 
\textbf{H} = \textbf{I} - \widetilde{\textbf{L}},
\label{FILTER}
\end{equation}
where $\widetilde{\textbf{L}}$ denotes the symmetric normalized graph Laplacian matrix. Subsequently, we stack up $t$-layers graph Laplacian filters as follows:
\begin{equation}
\begin{aligned}
\textbf{X}_{s} &= (\prod_{i=1}^{t}\textbf{H})\textbf{X}
\\
&= \textbf{H}^t\textbf{X},
\label{SMOOTH}
\end{aligned}
\end{equation}
where $\textbf{X}_s$ denotes the smoothed attribute matrix. Besides, $\textbf{H}^t$ denotes the stacked $t$-layer graph Laplacian filters, which can filter out the high-frequency noise in node attributes.


Through this low-pass denoising operation, high-frequency noise in attributes are filtered out, thus improving the clustering performance and training efficiency (See Section \ref{low_pass_SCM_sec} and Section \ref{costs_sec}).

\subsection{Structural Contrastive Module}
In this section, we design the Structure Contrastive Module (SCM) to keep the structural consistency even across two different views, thus enhancing the discriminative capability of the network.


To be specific, we firstly encode the smoothed attributes $\textbf{X}_s$ with the designed parameter un-shared MLP encoders and then normalize the learned node embeddings with ${\ell ^2}$-norm as follows:
\begin{equation} 
\begin{aligned}
\textbf{Z}^{v_1} &= \text{MLP}_1(\textbf{X}_s), \textbf{Z}^{v_1} = \frac{\textbf{Z}^{v_1}}{||\textbf{Z}^{v_1}||_2}, \\
\textbf{Z}^{v_2} &= \text{MLP}_2(\textbf{X}_s),  \textbf{Z}^{v_2} = \frac{\textbf{Z}^{v_2}}{||\textbf{Z}^{v_2}||_2},
\end{aligned}
\label{ENCODER}
\end{equation}
where $\textbf{Z}^{v_1}$ and $\textbf{Z}^{v_2}$ denote two augmented views of the learned node embeddings. It is worth to mentioning that $\text{MLP}_1$ and $\text{MLP}_2$ have the same architecture but un-shared parameters, thus $\textbf{Z}^{v_1}$ and $\textbf{Z}^{v_2}$ would contain different semantic information during training.


In addition, we further keep the difference between the two views by simply adding the random Gaussian noise to $\textbf{Z}^{v_2}$ as formulated:
\begin{equation} 
\textbf{Z}^{v_2} = \textbf{Z}^{v_2}+\textbf{N},
\label{NOISE}
\end{equation}
where $\textbf{N} \in \mathds{R}^{N \times d}$ is sampled from the Gaussian distribution $\mathcal{N}(0, \sigma)$. In summary, we construct two augmented views $\textbf{Z}^{v_1}$ and $\textbf{Z}^{v_2}$ by designing parameter un-shared encoders and perturbing the node embeddings directly instead of introducing complex operations against graphs, thus improving the training efficiency (See section \ref{costs_sec}). Besides, recent works \cite{MoCL,graph_aug1,AFGRL} have indicated that the complex data augmentations over graphs, like edge adding, edge dropping, and graph diffusion could lead to semantic drift. The similar conclusion is verified through experiments in Section \ref{data_aug_sec}.








Subsequently, we design a novel neighbor-oriented contrastive loss to keep cross-view structural consistency. Concretely, we calculate the cross-view sample similarity matrix $\textbf{S} \in \mathds{R}^{N \times N}$ between $\textbf{Z}^{v_1}$ and $\textbf{Z}^{v_2}$ as formulated:
\begin{equation}
\textbf{S}_{ij}= \textbf{Z}^{v_1}_i \cdot (\textbf{Z}^{v_2}_j)^{\text{T}}, \,\,  \forall\,\,i, j \in [1, N],
\label{CROSS_VIEW_MATRIX}
\end{equation}
where $\textbf{S}_{ij}$ denotes the cosine similarity between $i$-th node embedding in the first view and $j$-th node embedding in the second view. Then we force the cross-view sample similarity matrix $\textbf{S}$ to be equal to the the self-looped adjacency matrix $\widehat{\textbf{A}} \in \mathds{R}^{N\times N}$ as formulated: 
\begin{equation}
\begin{aligned}
\mathcal{L} &= \frac{1}{N^2}\sum (\textbf{S}-\widehat{\textbf{A}})^2 \\ &= \frac{1}{N^2}(\sum_i\sum_j\mathbbm{1}_{ij}^{1}(\textbf{S}_{ij}-1)^2+\sum_i\sum_j\mathbbm{1}_{ij}^{0}\textbf{S}_{ij}^2),
\end{aligned}
\label{SAMPLE_LOSS}
\end{equation}
where $\mathbbm{1}_{ij}^{1}$ denotes if $\widehat{\textbf{A}}_{ij}=1$ and $\mathbbm{1}_{ij}^{0}$ denotes if $\widehat{\textbf{A}}_{ij}=0$. Here, we consider the cross-view neighbors of the same node as the positive samples while regarding other non-neighbor nodes as negative samples. Then we pull together the positive samples while pushing away the negative samples. More precisely, in Eq. \eqref{SAMPLE_LOSS}, the first term forces the nodes to agree with their neighbors even across two different views while the second term minimizes the agreement between the node and its non-neighbors. This neighbor-oriented contrastive objective function enhances the discriminative capability of our network by keeping the cross-view structural consistency, thus improving the clustering performance (See Section \ref{low_pass_SCM_sec}).


\subsection{Fusion and Clustering}
In this section, we firstly fuse the two augmented views of the node embeddings in a linear manner as formulated:
\begin{equation}
\textbf{Z} = \frac{1}{2}(\textbf{Z}^{v_1}+\textbf{Z}^{v_2}),
\label{FUSION}
\end{equation}
where $\textbf{Z} \in \mathds{R}^{N \times d}$ denotes the resultant clustering-oriented node embeddings. Then we directly perform K-means algorithm \cite{K-means} over $\textbf{Z}$ and obtain the clustering results.

\begin{table*}[!t]
\caption{Performance comparison on seven datasets. All results are reported with mean±std under ten runs. The red and blue values indicate the best and the runner-up results, respectively. OOM indicates Out-Of-Memory during training.}
\scalebox{0.66}{
\begin{tabular}{c|c|cccccccccccccc}
\hline
{\color[HTML]{000000} }                                    & {\color[HTML]{000000} }                                  & {\color[HTML]{000000} \textbf{K-Means}} & {\color[HTML]{000000} \textbf{AE}} & {\color[HTML]{000000} \textbf{DEC}} & \textbf{SSGC} & {\color[HTML]{000000} \textbf{GAE}} & {\color[HTML]{000000} \textbf{DAEGC}} & {\color[HTML]{000000} \textbf{ARGA}} & {\color[HTML]{000000} \textbf{SDCN}} & {\color[HTML]{000000} \textbf{DFCN}} & \textbf{AGE}                      & {\color[HTML]{000000} \textbf{MVGRL}} & {\color[HTML]{000000} \textbf{SCAGC}}        & {\color[HTML]{000000} \textbf{MCGC}}         & \textbf{SCGC}                     \\ \cline{3-16} 
\multirow{-2}{*}{{\color[HTML]{000000} \textbf{Dataset}}}  & \multirow{-2}{*}{{\color[HTML]{000000} \textbf{Metric}}} & \cite{K-means}                                       & \cite{AE_K_MEANS}                                  & \cite{DEC}                                   & \cite{SSGC}             & \cite{GAE}                                   & \cite{DAEGC} & \cite{ARGA_conf}                                    & \cite{SDCN}                                    & \cite{DFCN}                                    & \cite{AGE}                                & \cite{MVGRL}                                    & \cite{SCAGC}                                           & \cite{MCGC}                                           & Ours                                \\ \hline
{\color[HTML]{000000} }                                    & {\color[HTML]{000000} ACC}                               & {\color[HTML]{000000} 33.80±2.71}       & {\color[HTML]{000000} 49.38±0.91}  & {\color[HTML]{000000} 46.50±0.26}   & 69.28±3.70    & {\color[HTML]{000000} 43.38±2.11}   & {\color[HTML]{000000} 70.43±0.36}     & {\color[HTML]{000000} 71.04±0.25}    & {\color[HTML]{000000} 35.60±2.83}    & {\color[HTML]{000000} 36.33±0.49}    & {\color[HTML]{0000FF} 73.50±1.83} & {\color[HTML]{000000} 70.47±3.70}     & {\color[HTML]{000000} 60.89±1.21}            & {\color[HTML]{000000} 42.85±1.13}            & {\color[HTML]{FE0000} 73.88±0.88} \\
{\color[HTML]{000000} }                                    & {\color[HTML]{000000} NMI}                               & {\color[HTML]{000000} 14.98±3.43}       & {\color[HTML]{000000} 25.65±0.65}  & {\color[HTML]{000000} 23.54±0.34}   & 54.32±1.92    & {\color[HTML]{000000} 28.78±2.97}   & {\color[HTML]{000000} 52.89±0.69}     & {\color[HTML]{000000} 51.06±0.52}    & {\color[HTML]{000000} 14.28±1.91}    & {\color[HTML]{000000} 19.36±0.87}    & {\color[HTML]{FF0000} 57.58±1.42} & {\color[HTML]{000000} 55.57±1.54}     & {\color[HTML]{000000} 39.72±0.72}            & {\color[HTML]{000000} 24.11±1.00}            & {\color[HTML]{0000FF} 56.10±0.72} \\
{\color[HTML]{000000} }                                    & {\color[HTML]{000000} ARI}                               & {\color[HTML]{000000} 08.60±1.95}       & {\color[HTML]{000000} 21.63±0.58}  & {\color[HTML]{000000} 15.13±0.42}   & 46.27±4.01    & {\color[HTML]{000000} 16.43±1.65}   & {\color[HTML]{000000} 49.63±0.43}     & {\color[HTML]{000000} 47.71±0.33}    & {\color[HTML]{000000} 07.78±3.24}    & {\color[HTML]{000000} 04.67±2.10}    & {\color[HTML]{0000FF} 50.60±2.14} & {\color[HTML]{000000} 48.70±3.94}     & {\color[HTML]{000000} 30.95±1.42}            & {\color[HTML]{000000} 14.33±1.26}            & {\color[HTML]{FE0000} 51.79±1.59} \\
\multirow{-4}{*}{{\color[HTML]{000000} \textbf{CORA}}}     & {\color[HTML]{000000} F1}                                & {\color[HTML]{000000} 30.26±4.46}       & {\color[HTML]{000000} 43.71±1.05}  & {\color[HTML]{000000} 39.23±0.17}   & 64.70±5.53    & {\color[HTML]{000000} 33.48±3.05}   & {\color[HTML]{000000} 68.27±0.57}     & {\color[HTML]{000000} 69.27±0.39}    & {\color[HTML]{000000} 24.37±1.04}    & {\color[HTML]{000000} 26.16±0.50}    & {\color[HTML]{0000FF} 69.68±1.59} & {\color[HTML]{000000} 67.15±1.86}     & {\color[HTML]{000000} 59.13±1.85}            & {\color[HTML]{000000} 35.16±0.91}            & {\color[HTML]{FE0000} 70.81±1.96} \\ \hline
{\color[HTML]{000000} }                                    & {\color[HTML]{000000} ACC}                               & {\color[HTML]{000000} 39.32±3.17}       & {\color[HTML]{000000} 57.08±0.13}  & {\color[HTML]{000000} 55.89±0.20}   & 68.97±0.34    & {\color[HTML]{000000} 61.35±0.80}   & {\color[HTML]{000000} 64.54±1.39}     & {\color[HTML]{000000} 61.07±0.49}    & {\color[HTML]{000000} 65.96±0.31}    & {\color[HTML]{000000} 69.50±0.20}    & {\color[HTML]{0000FF} 69.73±0.24} & {\color[HTML]{000000} 62.83±1.59}     & {\color[HTML]{000000} 61.16±0.72}            & {\color[HTML]{000000} 64.76±0.07}            & {\color[HTML]{FE0000} 71.02±0.77} \\
{\color[HTML]{000000} }                                    & {\color[HTML]{000000} NMI}                               & {\color[HTML]{000000} 16.94±3.22}       & {\color[HTML]{000000} 27.64±0.08}  & {\color[HTML]{000000} 28.34±0.30}   & 42.81±0.20    & {\color[HTML]{000000} 34.63±0.65}   & {\color[HTML]{000000} 36.41±0.86}     & {\color[HTML]{000000} 34.40±0.71}    & {\color[HTML]{000000} 38.71±0.32}    & {\color[HTML]{000000} 43.90±0.20}    & {\color[HTML]{0000FF} 44.93±0.53} & {\color[HTML]{000000} 40.69±0.93}     & {\color[HTML]{000000} 32.83±1.19}            & {\color[HTML]{000000} 39.11±0.06}            & {\color[HTML]{FE0000} 45.25±0.45} \\
{\color[HTML]{000000} }                                    & {\color[HTML]{000000} ARI}                               & {\color[HTML]{000000} 13.43±3.02}       & {\color[HTML]{000000} 29.31±0.14}  & {\color[HTML]{000000} 28.12±0.36}   & 44.42±0.32    & {\color[HTML]{000000} 33.55±1.18}   & {\color[HTML]{000000} 37.78±1.24}     & {\color[HTML]{000000} 34.32±0.70}    & {\color[HTML]{000000} 40.17±0.43}    & {\color[HTML]{0000FF} 45.50±0.30}    & {\color[HTML]{000000} 45.31±0.41} & {\color[HTML]{000000} 34.18±1.73}     & {\color[HTML]{000000} 31.17±0.23}            & {\color[HTML]{000000} 37.54±0.12}            & {\color[HTML]{FE0000} 46.29±1.13} \\
\multirow{-4}{*}{{\color[HTML]{000000} \textbf{CITESEER}}} & {\color[HTML]{000000} F1}                                & {\color[HTML]{000000} 36.08±3.53}       & {\color[HTML]{000000} 53.80±0.11}  & {\color[HTML]{000000} 52.62±0.17}   & 64.49±0.27    & {\color[HTML]{000000} 57.36±0.82}   & {\color[HTML]{000000} 62.20±1.32}     & {\color[HTML]{000000} 58.23±0.31}    & {\color[HTML]{000000} 63.62±0.24}    & {\color[HTML]{000000} 64.30±0.20}    & {\color[HTML]{0000FF} 64.45±0.27} & {\color[HTML]{000000} 59.54±2.17}     & {\color[HTML]{000000} 56.82±0.43}            & {\color[HTML]{000000} 59.64±0.05}            & {\color[HTML]{FE0000} 64.80±1.01} \\ \hline
{\color[HTML]{000000} }                                    & {\color[HTML]{000000} ACC}                               & {\color[HTML]{000000} 27.22±0.76}       & {\color[HTML]{000000} 48.25±0.08}  & {\color[HTML]{000000} 47.22±0.08}   & 60.23±0.19    & {\color[HTML]{000000} 71.57±2.48}   & {\color[HTML]{000000} 75.96±0.23}     & {\color[HTML]{000000} 69.28±2.30}    & {\color[HTML]{000000} 53.44±0.81}    & {\color[HTML]{0000FF} 76.82±0.23}    & {\color[HTML]{000000} 75.98±0.68} & {\color[HTML]{000000} 41.07±3.12}     & {\color[HTML]{000000} 75.25±0.10}            & {\color[HTML]{000000} }                      & {\color[HTML]{FE0000} 77.48±0.37} \\
{\color[HTML]{000000} }                                    & {\color[HTML]{000000} NMI}                               & {\color[HTML]{000000} 13.23±1.33}       & {\color[HTML]{000000} 38.76±0.30}  & {\color[HTML]{000000} 37.35±0.05}   & 60.37±0.15    & {\color[HTML]{000000} 62.13±2.79}   & {\color[HTML]{000000} 65.25±0.45}     & {\color[HTML]{000000} 58.36±2.76}    & {\color[HTML]{000000} 44.85±0.83}    & {\color[HTML]{000000} 66.23±1.21}    & {\color[HTML]{000000} 65.38±0.61} & {\color[HTML]{000000} 30.28±3.94}     & {\color[HTML]{0000FF} 67.18±0.13}            & {\color[HTML]{000000} }                      & {\color[HTML]{FE0000} 67.67±0.88} \\
{\color[HTML]{000000} }                                    & {\color[HTML]{000000} ARI}                               & {\color[HTML]{000000} 05.50±0.44}       & {\color[HTML]{000000} 20.80±0.47}  & {\color[HTML]{000000} 18.59±0.04}   & 35.99±0.47    & {\color[HTML]{000000} 48.82±4.57}   & {\color[HTML]{000000} 58.12±0.24}     & {\color[HTML]{000000} 44.18±4.41}    & {\color[HTML]{000000} 31.21±1.23}    & {\color[HTML]{0000FF} 58.28±0.74}    & {\color[HTML]{000000} 55.89±1.34} & {\color[HTML]{000000} 18.77±2.34}     & {\color[HTML]{000000} 56.86±0.23}            & {\color[HTML]{000000} }                      & {\color[HTML]{FE0000} 58.48±0.72} \\
\multirow{-4}{*}{{\color[HTML]{000000} \textbf{AMAP}}}     & {\color[HTML]{000000} F1}                                & {\color[HTML]{000000} 23.96±0.51}       & {\color[HTML]{000000} 47.87±0.20}  & {\color[HTML]{000000} 46.71±0.12}   & 52.79±0.01    & {\color[HTML]{000000} 68.08±1.76}   & {\color[HTML]{000000} 69.87±0.54}     & {\color[HTML]{000000} 64.30±1.95}    & {\color[HTML]{000000} 50.66±1.49}    & {\color[HTML]{000000} 71.25±0.31}    & {\color[HTML]{000000} 71.74±0.93} & {\color[HTML]{000000} 32.88±5.50}     & {\color[HTML]{FF0000} 72.77±0.16}            & \multirow{-4}{*}{{\color[HTML]{000000} OOM}} & {\color[HTML]{0000FF} 72.22±0.97} \\ \hline
{\color[HTML]{000000} }                                    & {\color[HTML]{000000} ACC}                               & {\color[HTML]{000000} 40.23±1.19}       & {\color[HTML]{000000} 47.79±3.95}  & {\color[HTML]{000000} 42.09±2.21}   & 36.11±2.16    & {\color[HTML]{000000} 53.59±2.04}   & {\color[HTML]{000000} 52.67±0.00}     & {\color[HTML]{0000FF} 67.86±0.80}    & {\color[HTML]{000000} 53.05±4.63}    & {\color[HTML]{000000} 55.73±0.06}    & {\color[HTML]{000000} 56.68±0.76} & {\color[HTML]{000000} 37.56±0.32}     & {\color[HTML]{000000} 57.25±1.65}            & {\color[HTML]{000000} 38.93±0.23}            & {\color[HTML]{FE0000} 77.97±0.99} \\
{\color[HTML]{000000} }                                    & {\color[HTML]{000000} NMI}                               & {\color[HTML]{000000} 26.92±2.39}       & {\color[HTML]{000000} 18.03±7.73}  & {\color[HTML]{000000} 14.10±1.99}   & 13.74±1.60    & {\color[HTML]{000000} 30.59±2.06}   & {\color[HTML]{000000} 21.43±0.35}     & {\color[HTML]{0000FF} 49.09±0.54}    & {\color[HTML]{000000} 25.74±5.71}    & {\color[HTML]{000000} 48.77±0.51}    & {\color[HTML]{000000} 36.04±1.54} & {\color[HTML]{000000} 29.33±0.70}     & {\color[HTML]{000000} 22.18±0.31}            & {\color[HTML]{000000} 23.11±0.56}            & {\color[HTML]{FE0000} 52.91±0.68} \\
{\color[HTML]{000000} }                                    & {\color[HTML]{000000} ARI}                               & {\color[HTML]{000000} 09.52±1.42}       & {\color[HTML]{000000} 13.75±6.05}  & {\color[HTML]{000000} 07.99±1.21}   & 4.00±1.98     & {\color[HTML]{000000} 24.15±1.70}   & {\color[HTML]{000000} 18.18±0.29}     & {\color[HTML]{0000FF} 42.02±1.21}    & {\color[HTML]{000000} 21.04±4.97}    & {\color[HTML]{000000} 37.76±0.23}    & {\color[HTML]{000000} 26.59±1.83} & {\color[HTML]{000000} 13.45±0.03}     & {\color[HTML]{000000} 27.29±1.53}            & {\color[HTML]{000000} 08.41±0.32}            & {\color[HTML]{FE0000} 50.64±1.85} \\
\multirow{-4}{*}{{\color[HTML]{000000} \textbf{BAT}}}      & {\color[HTML]{000000} F1}                                & {\color[HTML]{000000} 34.45±2.10}       & {\color[HTML]{000000} 46.80±3.44}  & {\color[HTML]{000000} 42.63±2.35}   & 29.74±2.76    & {\color[HTML]{000000} 50.83±3.23}   & {\color[HTML]{000000} 52.23±0.03}     & {\color[HTML]{0000FF} 67.02±1.15}    & {\color[HTML]{000000} 46.45±5.90}    & {\color[HTML]{000000} 50.90±0.12}    & {\color[HTML]{000000} 55.07±0.80} & {\color[HTML]{000000} 29.64±0.49}     & {\color[HTML]{000000} 52.53±0.54}            & {\color[HTML]{000000} 32.92±0.25}            & {\color[HTML]{FE0000} 78.03±0.96} \\ \hline
{\color[HTML]{000000} }                                    & {\color[HTML]{000000} ACC}                               & {\color[HTML]{000000} 32.23±0.56}       & {\color[HTML]{000000} 38.85±2.32}  & {\color[HTML]{000000} 36.47±1.60}   & 32.41±0.45    & {\color[HTML]{000000} 44.61±2.10}   & {\color[HTML]{000000} 36.89±0.15}     & {\color[HTML]{0000FF} 52.13±0.00}    & {\color[HTML]{000000} 39.07±1.51}    & {\color[HTML]{000000} 49.37±0.19}    & {\color[HTML]{000000} 47.26±0.32} & {\color[HTML]{000000} 32.88±0.71}     & {\color[HTML]{000000} 44.61±1.57}            & {\color[HTML]{000000} 32.58±0.29}            & {\color[HTML]{FF0000} 57.94±0.42} \\
{\color[HTML]{000000} }                                    & NMI                                                      & 11.02±1.21                              & 06.92±2.80                         & 04.96±1.74                          & 4.65±0.21     & 15.60±2.30                          & 05.57±0.06                            & 22.48±1.21                           & 08.83±2.54                           & {\color[HTML]{0000FF} 32.90±0.41}    & 23.74±0.90                        & {\color[HTML]{000000} 11.72±1.08}     & {\color[HTML]{000000} 07.32±1.97}            & {\color[HTML]{000000} 07.04±0.56}            & {\color[HTML]{FF0000} 33.91±0.49} \\
{\color[HTML]{000000} }                                    & ARI                                                      & 02.20±0.40                              & 05.11±2.65                         & 03.60±1.87                          & 1.53±0.04     & 13.40±1.26                          & 05.03±0.08                            & 17.29±0.50                           & 06.31±1.95                           & {\color[HTML]{0000FF} 23.25±0.18}    & 16.57±0.46                        & {\color[HTML]{000000} 04.68±1.30}     & {\color[HTML]{000000} 11.33±1.47}            & {\color[HTML]{000000} 01.33±0.14}            & {\color[HTML]{FF0000} 27.51±0.59} \\
\multirow{-4}{*}{{\color[HTML]{000000} \textbf{EAT}}}      & F1                                                       & 23.49±0.92                              & 38.75±2.25                         & 34.84±1.28                          & 26.49±0.66    & 43.08±3.26                          & 34.72±0.16                            & {\color[HTML]{0000FF} 52.75±0.07}    & {\color[HTML]{000000} 33.42±3.10}    & {\color[HTML]{000000} 42.95±0.04}    & {\color[HTML]{000000} 45.54±0.40} & {\color[HTML]{000000} 25.35±0.75}     & {\color[HTML]{000000} 44.14±0.24}            & {\color[HTML]{000000} 27.03±0.16}            & {\color[HTML]{FF0000} 57.96±0.46} \\ \hline
                                                           & ACC                                                      & 42.47±0.15                              & 46.82±1.14                         & 45.61±1.84                          & 36.74±0.81    & 48.97±1.52                          & 52.29±0.49                            & {\color[HTML]{000000} 49.31±0.15}    & {\color[HTML]{000000} 52.25±1.91}    & {\color[HTML]{000000} 33.61±0.09}    & {\color[HTML]{0000FF} 52.37±0.42} & {\color[HTML]{000000} 44.16±1.38}     & {\color[HTML]{000000} 50.75±0.64}            & {\color[HTML]{000000} 41.93±0.56}            & {\color[HTML]{FF0000} 56.58±1.62} \\
                                                           & {\color[HTML]{000000} NMI}                               & {\color[HTML]{000000} 22.39±0.69}       & {\color[HTML]{000000} 17.18±1.60}  & {\color[HTML]{000000} 16.63±2.39}   & 8.04±0.18     & {\color[HTML]{000000} 20.69±0.98}   & {\color[HTML]{000000} 21.33±0.44}     & {\color[HTML]{0000FF} 25.44±0.31}    & {\color[HTML]{000000} 21.61±1.26}    & {\color[HTML]{000000} 26.49±0.41}    & {\color[HTML]{000000} 23.64±0.66} & {\color[HTML]{000000} 21.53±0.94}     & {\color[HTML]{000000} 23.60±1.78}            & {\color[HTML]{000000} 16.64±0.41}            & {\color[HTML]{FF0000} 28.07±0.71} \\
                                                           & {\color[HTML]{000000} ARI}                               & {\color[HTML]{000000} 15.71±0.76}       & {\color[HTML]{000000} 13.59±2.02}  & {\color[HTML]{000000} 13.14±1.97}   & 5.12±0.27     & {\color[HTML]{000000} 18.33±1.79}   & {\color[HTML]{000000} 20.50±0.51}     & {\color[HTML]{000000} 16.57±0.31}    & {\color[HTML]{000000} 21.63±1.49}    & {\color[HTML]{000000} 11.87±0.23}    & {\color[HTML]{000000} 20.39±0.70} & {\color[HTML]{000000} 17.12±1.46}     & {\color[HTML]{0000FF} 23.33±0.32}            & {\color[HTML]{000000} 12.21±0.13}            & {\color[HTML]{FE0000} 24.80±1.85} \\
\multirow{-4}{*}{\textbf{UAT}}                             & {\color[HTML]{000000} F1}                                & {\color[HTML]{000000} 36.12±0.22}       & {\color[HTML]{000000} 45.66±1.49}  & {\color[HTML]{000000} 44.22±1.51}   & 29.50±1.57    & {\color[HTML]{000000} 47.95±1.52}   & {\color[HTML]{0000FF} 50.33±0.64}     & {\color[HTML]{000000} 50.26±0.16}    & {\color[HTML]{000000} 45.59±3.54}    & {\color[HTML]{000000} 25.79±0.29}    & {\color[HTML]{000000} 50.15±0.73} & {\color[HTML]{000000} 39.44±2.19}     & {\color[HTML]{000000} 47.07±0.73}            & {\color[HTML]{000000} 35.78±0.38}            & {\color[HTML]{FE0000} 55.52±0.87} \\ \hline
{\color[HTML]{000000} }                                    & {\color[HTML]{000000} ACC}                               & {\color[HTML]{000000} 26.27±1.10}       & {\color[HTML]{000000} 33.12±0.19}  & {\color[HTML]{000000} 31.92±0.45}   & 37.43±0.55    & {\color[HTML]{000000} 29.60±0.81}   & {\color[HTML]{000000} 34.35±1.00}     & {\color[HTML]{000000} 22.07±0.43}    & {\color[HTML]{000000} 26.67±0.40}    & {\color[HTML]{000000} 37.51±0.81}    & {\color[HTML]{0000FF} 39.62±0.13} & {\color[HTML]{000000} 31.52±2.95}     & {\color[HTML]{000000} }                      & {\color[HTML]{000000} 29.08±0.58}            & {\color[HTML]{FE0000} 41.89±0.47} \\
{\color[HTML]{000000} }                                    & {\color[HTML]{000000} NMI}                               & {\color[HTML]{000000} 34.68±0.84}       & {\color[HTML]{000000} 41.53±0.25}  & {\color[HTML]{000000} 41.67±0.24}   & 53.74±0.21    & {\color[HTML]{000000} 45.82±0.75}   & {\color[HTML]{000000} 49.16±0.73}     & {\color[HTML]{000000} 41.28±0.25}    & {\color[HTML]{000000} 37.38±0.39}    & {\color[HTML]{000000} 51.30±0.41}    & {\color[HTML]{0000FF} 52.38±0.17} & {\color[HTML]{000000} 48.99±3.95}     & {\color[HTML]{000000} }                      & {\color[HTML]{000000} 36.86±0.56}            & {\color[HTML]{FE0000} 53.00±0.29} \\
{\color[HTML]{000000} }                                    & {\color[HTML]{000000} ARI}                               & {\color[HTML]{000000} 09.35±0.57}       & {\color[HTML]{000000} 18.13±0.27}  & {\color[HTML]{000000} 16.98±0.29}   & 26.23±0.26    & {\color[HTML]{000000} 17.84±0.86}   & {\color[HTML]{000000} 22.60±0.47}     & {\color[HTML]{000000} 12.38±0.24}    & {\color[HTML]{000000} 13.63±0.27}    & {\color[HTML]{000000} 24.46±0.48}    & {\color[HTML]{FF0000} 24.46±0.23} & {\color[HTML]{000000} 19.11±2.63}     & {\color[HTML]{000000} }                      & {\color[HTML]{000000} 13.15±0.48}            & {\color[HTML]{0000FF} 24.23±0.86} \\
\multirow{-4}{*}{{\color[HTML]{000000} \textbf{CORAFULL}}} & {\color[HTML]{000000} F1}                                & {\color[HTML]{000000} 22.57±1.09}       & {\color[HTML]{000000} 28.40±0.30}  & {\color[HTML]{000000} 27.71±0.58}   & 29.50±0.88    & {\color[HTML]{000000} 25.95±0.75}   & {\color[HTML]{000000} 26.96±1.33}     & {\color[HTML]{000000} 18.85±0.41}    & {\color[HTML]{000000} 22.14±0.43}    & {\color[HTML]{0000FF} 31.22±0.87}    & {\color[HTML]{000000} 27.95±0.19} & {\color[HTML]{000000} 26.51±2.87}     & \multirow{-4}{*}{{\color[HTML]{000000} OOM}} & {\color[HTML]{000000} 22.90±0.52}            & {\color[HTML]{FE0000} 32.98±0.73} \\ \hline
\end{tabular}}
\label{COMPARE_TABLE}
\end{table*}

\begin{algorithm}[h]
\small
\caption{Simple Contrastive Graph Clustering (SCGC)}
\label{ALGORITHM}
\flushleft{\textbf{Input}: The input graph $\mathcal{G}=\{\textbf{X},\textbf{A}\}$; The cluster number $C$; The iteration number $I$; The graph Laplacian filter layer number $t$; The Gaussian noise's standard deviation $\sigma$.} \\
\flushleft{\textbf{Output}: The clustering result \textbf{R}.} 
\begin{algorithmic}[1]
\STATE Obtain the smoothed attributes $\textbf{X}_s$ by applying $t$-layers stacked graph Laplacian filters to the attributes $\textbf{X}$ in Eq. \eqref{FILTER}-\eqref{SMOOTH}.
\FOR{$i=1$ to $I$}
\STATE Encode $\textbf{X}_s$ into two augmented views with parameter un-shared siamese MLP encoders and then normalize them in Eq. \eqref{ENCODER}.
\STATE Perturb node embeddings by adding the Gaussian noise in Eq. \eqref{NOISE}.
\STATE Calculate cross-view sample similarity matrix $\textbf{S}$ by Eq. \eqref{CROSS_VIEW_MATRIX}.
\STATE Force $\textbf{S}$ to approach the self-looped adjacency $\widehat{\textbf{A}}$ and calculate the neighbor-oriented contrastive loss ${\mathcal{L}}$ in Eq. \eqref{SAMPLE_LOSS}.
\STATE Fuse $\textbf{Z}^{v_1}$ and $\textbf{Z}^{v_2}$ to obtain $\textbf{Z}$ in Eq. \eqref{FUSION}.
\STATE Update model by minimizing $\mathcal{L}$ with Adam optimizer.
\ENDFOR
\STATE{Obtain \textbf{R} by performing K-means over $\textbf{Z}$.}
\STATE \textbf{return} \textbf{R}
\end{algorithmic}
\end{algorithm}

\subsection{Objective Function}
The optimization objective of the proposed method is the neighbor-oriented contrastive loss $\mathcal{L}$ in Eq. \eqref{SAMPLE_LOSS}. We minimize $\mathcal{L}$ with the widely-used Adam optimizer \cite{ADAM} during training. The detailed learning process of our proposed SCGC is shown in Algorithm \ref{ALGORITHM}.

\section{EXPERIMENT}
\subsection{Dataset}
To evaluate the effectiveness and efficiency of our proposed SCGC, we conduct extensive experiments on seven benchmark datasets, including CORA \cite{AGE}, CITESEER \cite{AGE}, Brazil Air-Traffic (BAT) \cite{RETHINK}, Europe Air-Traffic (EAT) \cite{RETHINK}, USA Air-Traffic (UAT) \cite{RETHINK}, Amazon Photo (AMAP) \cite{DCRN}, and CORAFULL \cite{DCRN}. The brief information of these datasets is summarized in Table \ref{DATASET_INFO}.

\begin{table}[h]
\centering
\caption{Statistics summary of seven datasets.}
\small
\scalebox{0.95}{
\begin{tabular}{@{}cccccc@{}}
\toprule
\textbf{Dataset} & \textbf{Type} & \textbf{Sample} & \textbf{Dimension} & \textbf{Edge}  & \textbf{Class} \\ \midrule
\textbf{CORA}  & Graph   & 2708    & 1433       & 5429   & 7       \\
\textbf{CITESEER} & Graph    & 3327    & 3703      & 4732   & 6       \\
\textbf{AMAP} & Graph  & 7650   & 745       & 119081  & 8       \\
\textbf{BAT}    & Graph  & 131    & 81      & 1038  & 4       \\
\textbf{EAT}    & Graph & 399    & 203       & 5994 & 4       \\
\textbf{UAT} & Graph  & 1190   & 239       & 13599  & 4       \\
\textbf{CORAFULL} & Graph & 19793   & 8710      & 63421 & 70      \\ \bottomrule
\end{tabular}}
\label{DATASET_INFO} 
\end{table}

\subsection{Experiment Setup}
All experimental results are obtained from the desktop computer with the Intel Core i7-6800K CPU, one NVIDIA GeForce RTX 3090 GPU, 64GB RAM, and the PyTorch deep learning platform.

\subsubsection{\textbf{Training Procedure}}
Our network is trained for 400 epochs until convergence by minimizing the contrastive loss in Eq. \eqref{SAMPLE_LOSS} with Adam optimizer \cite{ADAM}. After optimization, we directly perform K-means algorithm \cite{K-means} on the clustering-oriented node embeddings $\textbf{Z}$. To avoid the influence of randomness, we conduct ten runs for all compared methods and report the average values with standard deviations of four metrics.


\subsubsection{\textbf{Parameter Settings}}
To MCGC \cite{MCGC}, we run their source code on merely the graph datasets in Table \ref{DATASET_INFO} for fairness. For other baselines, we reproduce results by adopting their source code with the original settings. In our proposed method, two MLPs both consist of a single 500-dimensional embedding layer. The learning rate of the optimizer is set to 1e-3 for CORA / BAT / EAT / UAT, 1e-4 for CORAFULL, 5e-5 for CITESEER, and 1e-5 for AMAP, respectively. The layer number $t$ of graph Laplacian filters is set to 2 for CORA / CITESEER / CORAFULL, 3 for BAT / UAT, and 5 for AMAP / EAT. The standard deviation $\sigma$ of random Gaussian noise is set to 0.01.

\subsubsection{\textbf{Metrics}} 
To verify the superiority of our SCGC compared with baselines, the clustering performance is evaluated by four widely used metrics, i.e., ACC, NMI, ARI, and F1 \cite{ZHOU_1,siwei_1,siwei_2}.

\subsection{Performance Comparison}
To demonstrate the superiority of our proposed Simple Contrastive Graph Clustering (SCGC) algorithm, we compare SCGC with thirteen baselines. Specifically, 
K-means \cite{K-means} is a classic clustering algorithm. Besides, two representative deep clustering methods, i.e., AE \cite{AE_K_MEANS} and DEC \cite{DEC}, encode nodes with auto-encoders and then perform the clustering algorithm over the learned embeddings. Besides, a simple spectral method SSGC \cite{SSGC} is designed to trade-off between low-pass and high-pass filter bands. In addition, five classical deep graph clustering methods \cite{GAE,DAEGC,ARGA_conf,SDCN,DFCN} utilize the graph auto-encoder \cite{GAE} to learn the node representation for clustering. Moreover, we test the clustering performance of four state-of-the-art contrastive deep graph clustering methods including AGE \cite{AGE}, MVGRL \cite{MVGRL}, SCAGC \cite{SCAGC}, and MCGC \cite{MCGC}, which design contrastive strategies to improve the discriminative capability of samples.

\begin{table*}[!t]
\centering
\caption{Time cost comparisons of the training process. All results are measured in seconds. The bold and underlined values indicate the best and the runner-up results, respectively. Avg. indicates the average time cost on six datasets. OOM denotes Out-Of-Memory during training.}
\scalebox{1.67}{
\tiny
\begin{tabular}{c|ccccc|ccccc}
\hline
\multirow{3}{*}{Method} & \multicolumn{5}{c|}{Classical Method}             & \multicolumn{5}{c}{Contrastive Method}                    \\ \cline{2-11} 
                        & DEC    & SSGC       & GAE         & DAEGC & SDCN  & AGE    & MVGRL      & MCGC       & SCAGC  & SCGC          \\
                        & \cite{DEC}      & \cite{SSGC}          & \cite{GAE}           & \cite{DAEGC}     & \cite{SDCN}     & \cite{AGE}      & \cite{MVGRL}          & \cite{MCGC}          & \cite{SCAGC}      & Ours            \\ \hline
CORA                    & 91.13  & 16.23      & {\underline{7.38}}  & 12.97 & 11.32 & 46.65  & 14.72      & 118.07     & 54.08  & \textbf{3.86} \\
CITESEER                & 223.95 & 51.29      & {\underline{6.69}}  & 14.70 & 11.00 & 70.63  & 18.31      & 126.06     & 50.00  & \textbf{4.88} \\
AMAP                    & 264.20 & 100.54     & {\underline{18.64}} & 39.62 & 19.28 & 377.49 & 131.38     & OOM        & 150.54 & \textbf{8.86} \\
BAT                     & 21.37  & 8.17       & 3.83        & 4.79  & 11.50 & 2.49   & 3.19       & {\underline{2.28}} & 93.79  & \textbf{1.40} \\
EAT                     & 26.99  & {\underline{2.52}} & 4.64        & 5.14  & 12.12 & 3.86   & 3.32       & 2.87       & 47.79  & \textbf{1.61} \\
UAT                     & 42.30  & 17.53      & 4.75        & 6.44  & 10.64 & 8.95   & {\underline{4.27}} & 23.10      & 64.70  & \textbf{1.68} \\
Avg.                    & 111.66 & 32.71      & {\underline{7.66}}  & 13.94 & 12.64 & 85.01  & 29.20      & 54.48      & 76.82  & \textbf{3.71} \\ \hline
\end{tabular}}
\label{TIME_TABLE}
\end{table*}

In Table \ref{COMPARE_TABLE}, we report the clustering performance of all compared methods on seven datasets. From these results, we have four observations as follows. 1) Since K-means is directly performed on the raw attributes, thus achieving unpromising results. 2) The spectral-based method SSGC is not comparable with ours since they overlook the strong supervision information exploitation capability of the contrastive learning. 3) Our SCGC exceeds the representative deep clustering methods \cite{AE_K_MEANS,DEC,IDEC} since they merely consider the node attributes while overlooking the topological information in graphs. 4) The recent contrastive methods \cite{AGE,MVGRL,MCGC,SCAGC} achieve sub-optimal performance compared with our proposed SCGC. The reason is that we improve the discriminative capability of samples by keep the cross-view structural consistency in the proposed structural contrastive module. 5) Taking the average value of four metrics into account, SCGC consistently outperforms all baselines on seven datasets. For example, on BAT dataset, SCGC exceeds the runner-up ARGA \cite{ARGA_conf} by 10.11\% 3.82\%, 8.62\% 11.01\% increments with respect to ACC, NMI, ARI, and F1.

Overall, the aforementioned observations have demonstrated the superiority of our proposed SCGC. In the following section, we will conduct experiments to verify the efficiency of SCGC.


\begin{figure}[!t]
\centering
\includegraphics[scale=0.42]{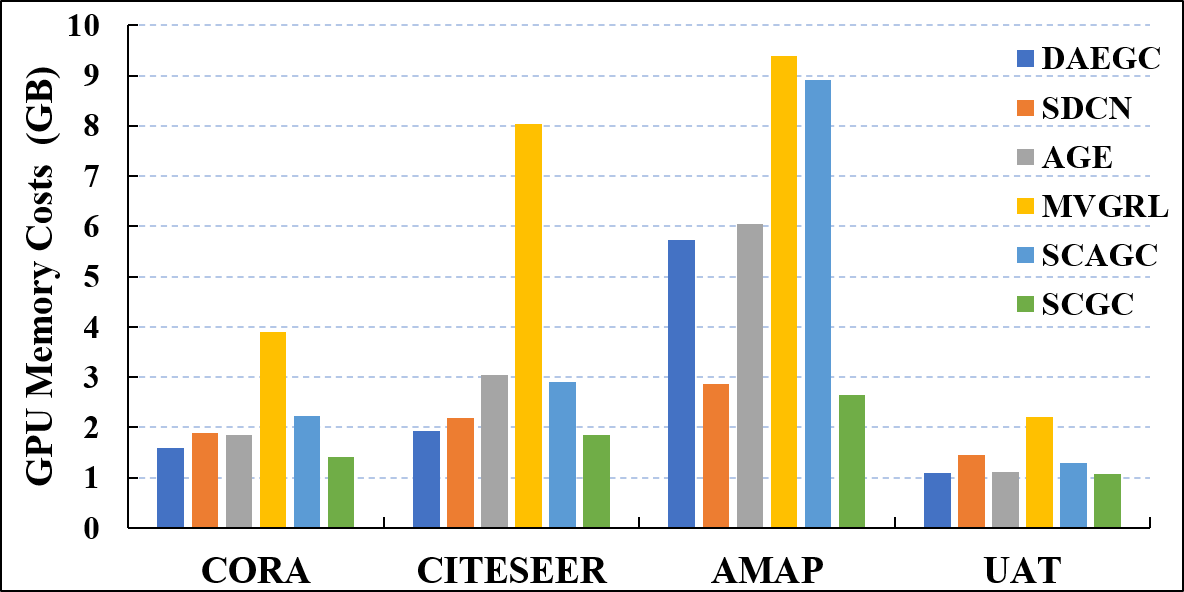}
\caption{GPU memory costs of six methods on four datasets.}
\label{GPU_Memory}  
\end{figure}

\subsection{Time Costs \& GPU Memory Costs} \label{costs_sec}
Time costs and GPU memory costs are two important indicators for evaluating the efficiency of algorithms. In this section, we conduct expensive experiments to demonstrate the efficiency of SCGC.

Firstly, we test the training time of our SCGC and nine baselines on six datasets. Concretely, the baselines contain one deep clustering method DEC \cite{DEC}, one spectral-based method SSGC \cite{SSGC}, three classical deep graph clustering methods \cite{GAE,DAEGC,SDCN}, and four contrastive methods \cite{AGE,MVGRL,SCAGC,MCGC}. For fairness, we train all methods for 400 epochs. From the results in Table \ref{TIME_TABLE}, we observe that our method consistently achieves the fastest speed on six datasets. Significantly, SCGC outperforms the recent contrastive deep clustering competitors with at least seven times speedup on average. We summarize two key reasons as follows. 1) The network architecture of SCGC is simple and merely consists of two MLPs. 2) Similar to \cite{AGE,SGC,graph_mlp,SSGC}, our method decouples the GCN \cite{GCN} and adopts the low-pass denoising operations as an independent pre-processing to conduct neighbor information aggregation, thus simplifying the training process.

\begin{table}[!t]
\small
\centering
\caption{Ablation studies of low-pass denoising operation and structural contrastive module. Results are reported with mean±std under ten runs. Bold values are the best results.}
\scalebox{0.75}{
\begin{tabular}{c|c|cccc}
\hline
\multirow{2}{*}{\textbf{Dataset}}  & \multirow{2}{*}{\textbf{Metric}} & \multirow{2}{*}{\textbf{(w/o) L \& SCM}} & \multirow{2}{*}{\textbf{(w/o) SCM}} & \multirow{2}{*}{\textbf{(w/o) L}} & \multirow{2}{*}{\textbf{L+SCM}} \\
                                   &                                  &                                    &                                   &                                   &                                  \\ \hline
\multirow{4}{*}{\textbf{CORA}}     & ACC                              & 33.80±2.71                         & 56.42±4.96                        & 57.81±0.82                        & \textbf{73.88±0.88}              \\
                                   & NMI                              & 14.98±3.43                         & 44.51±3.69                        & 34.65±1.25                        & \textbf{56.10±0.72}              \\
                                   & ARI                              & 08.60±1.95                         & 32.70±4.22                        & 28.70±1.27                        & \textbf{51.79±1.59}              \\
                                   & F1                               & 30.26±4.46                         & 47.75±7.31                        & 52.29±0.90                        & \textbf{70.81±1.96}              \\ \hline
\multirow{4}{*}{\textbf{CITESEER}} & ACC                              & 39.32±3.17                         & 57.48±3.94                        & 65.59±0.86                        & \textbf{71.02±0.77}              \\
                                   & NMI                              & 16.94±3.22                         & 37.95±2.30                        & 39.03±0.66                        & \textbf{45.25±0.45}              \\
                                   & ARI                              & 13.43±3.02                         & 34.00±2.76                        & 38.66±0.50                        & \textbf{46.29±1.13}              \\
                                   & F1                               & 36.08±3.53                         & 44.90±3.97                        & 60.75±1.25                        & \textbf{64.80±1.01}              \\ \hline
\multirow{4}{*}{\textbf{AMAP}}     & ACC                              & 27.22±0.76                         & 29.58±1.85                        & 45.32±1.69                        & \textbf{77.48±0.37}              \\
                                   & NMI                              & 13.23±1.33                         & 14.87±3.08                        & 29.58±2.98                        & \textbf{67.67±0.88}              \\
                                   & ARI                              & 05.50±0.44                         & 05.47±1.35                        & 19.26±2.02                        & \textbf{58.48±0.72}              \\
                                   & F1                               & 23.96±0.51                         & 26.01±2.88                        & 41.59±0.36                        & \textbf{72.22±0.97}              \\ \hline
\multirow{4}{*}{\textbf{BAT}}      & ACC                              & 40.23±1.19                         & 56.11±2.44                        & 70.46±0.38                        & \textbf{77.97±0.99}              \\
                                   & NMI                              & 26.92±2.39                         & 34.79±1.93                        & 48.47±0.37                        & \textbf{52.91±0.68}              \\
                                   & ARI                              & 9.520±1.42                         & 24.13±2.27                        & 45.11±0.36                        & \textbf{50.64±1.85}              \\
                                   & F1                               & 34.45±2.10                         & 54.35±3.13                        & 68.47±0.52                        & \textbf{78.03±0.96}              \\ \hline
\multirow{4}{*}{\textbf{EAT}}      & ACC                              & 32.23±0.56                         & 51.03±2.03                        & 53.86±1.12                        & \textbf{57.94±0.42}              \\
                                   & NMI                              & 11.02±1.21                         & 32.27±0.71                        & 29.17±1.63                        & \textbf{33.91±0.49}              \\
                                   & ARI                              & 02.20±0.40                         & 24.03±1.07                        & 23.87±2.24                        & \textbf{27.51±0.59}              \\
                                   & F1                               & 23.49±0.92                         & 47.99±1.74                        & 52.70±0.17                        & \textbf{57.96±0.46}              \\ \hline
\multirow{4}{*}{\textbf{UAT}}      & ACC                              & 42.47±0.15                         & 39.79±2.50                        & 46.45±0.79                        & \textbf{56.58±1.62}              \\
                                   & NMI                              & 22.39±0.69                         & 12.99±1.70                        & 21.66±0.94                        & \textbf{28.07±0.71}              \\
                                   & ARI                              & 15.71±0.76                         & 07.04±1.15                        & 15.35±0.64                        & \textbf{24.80±1.85}              \\
                                   & F1                               & 36.12±0.22                         & 38.09±4.19                        & 43.72±1.24                        & \textbf{55.52±0.87}              \\ \hline
\end{tabular}}
\label{structure_ablation}
\end{table}

Secondly, we conduct experiments to test GPU memory costs of SCGC and five baselines including two classical deep graph clustering methods \cite{DAEGC,SDCN} and three contrastive methods \cite{AGE,MVGRL,SCAGC} on four datasets. From the results in Fig. \ref{GPU_Memory}, two conclusions are obtained as follows. 1) SCGC achieves comparable memory costs as the classical deep graph clustering methods including DAEGC \cite{DAEGC} and SDCN \cite{SDCN}. 2) Compared to the contrastive methods \cite{AGE,MVGRL,SCAGC}, our proposed method saves about 59\% GPU memory on average. We summarize two reasons as follows. 1) The siamese MLP encoders in SCGC are light. 2) Our proposed method merely applies data augmentations in the latent space instead of introducing complex space-consuming operations over graphs \cite{MVGRL,SCAGC}.

\subsection{Ablation Studies}
\subsubsection{\textbf{Effectiveness of Low-pass Denoising Operation \& Structural Contrastive Module}} \label{low_pass_SCM_sec}
In this section, we conduct the ablation studies to verify the effectiveness of two components in our network, i.e., low-pass denoising operation and Structural Contrastive Module (SCM). Here, we denote the low-pass denoising operation as L for short. In Table \ref{structure_ablation}, ``L+SCM'' denotes our proposed SCGC. Besides, ``(w/o) L'', ``(w/o) SCM'', and ``(w/o) L \& SCM'' denote SCGC without L, SCM, and both of them, respectively. From these results, we have three observations as follows. 1) The low-pass denoising operation could improve the performance of the baseline by filtering out the high-frequency noise in node attributes. Concretely, the loss-pass denoising  ''L+SCM'' outperforms ''(w/o) L'' average 10.77\% ACC on six datasets. 2) Through our proposed SCM, the discriminative of samples is enhanced, thus achieving better performance compared to the baseline. We observe that the ''L+SCM'' outperforms ''(w/o) SCM'' average 17.78\% ACC 3) Our method consistently outperforms other variants by a large margin. Overall, the aforementioned observations have verified the effectiveness of the low-pass denoising operation and structural contrastive module in our proposed SCGC.

\begin{figure}[h]
\centering
\footnotesize
\begin{minipage}{0.48\linewidth}
\centerline{\includegraphics[width=0.90\textwidth]{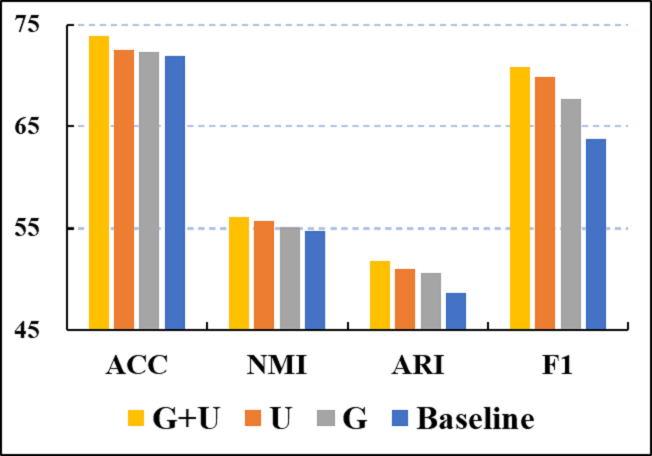}}
\vspace{3pt}
\centerline{CORA}
\vspace{3pt}
\centerline{\includegraphics[width=0.90\textwidth]{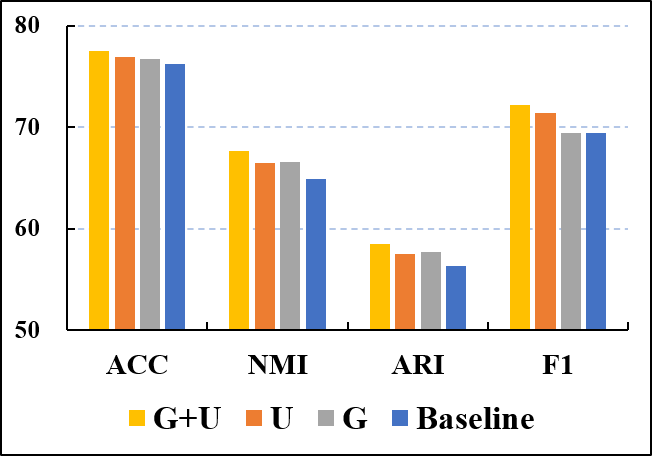}}
\vspace{3pt}
\centerline{AMAP}
\vspace{3pt}
\centerline{\includegraphics[width=0.90\textwidth]{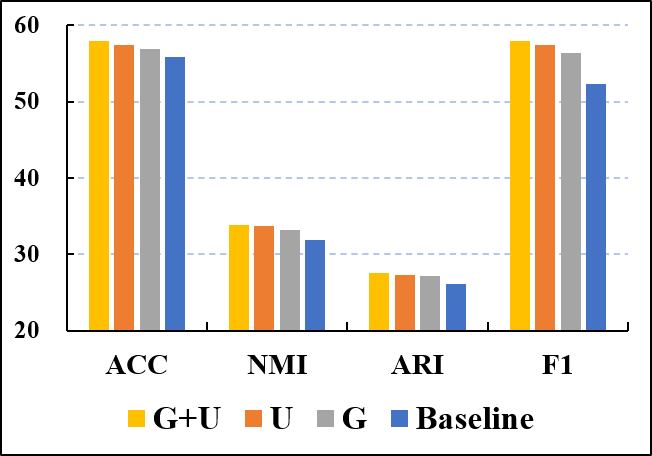}}
\vspace{3pt}
\centerline{EAT}
\end{minipage}
\begin{minipage}{0.48\linewidth}
\centerline{\includegraphics[width=0.90\textwidth]{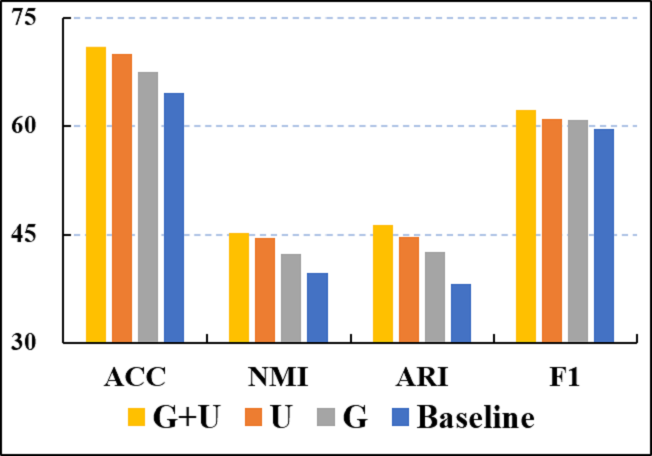}}
\vspace{3pt}
\centerline{CITESEER}
\vspace{3pt}
\centerline{\includegraphics[width=0.90\textwidth]{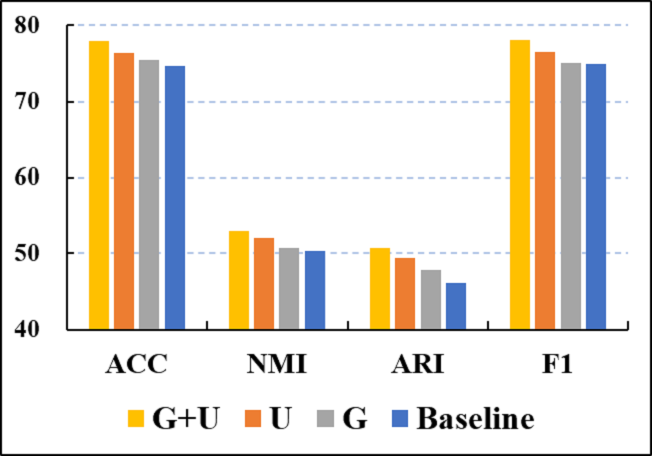}}
\vspace{3pt}
\centerline{BAT}
\vspace{3pt}
\centerline{\includegraphics[width=0.90\textwidth]{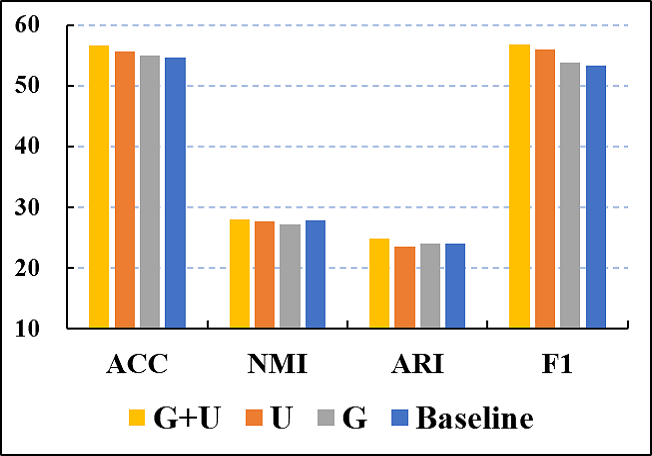}}
\vspace{3pt}
\centerline{UAT}
\end{minipage}
\caption{Ablation studies of un-shared MLPs and Gaussian noise on six datasets.}
\label{mlp_noise_ablation}
\end{figure}

\subsubsection{\textbf{Effectiveness of the proposed Data Augmentation}}\label{data_aug_sec}
In our proposed SCGC, we construct augmented views of the same node by designing parameter un-shared siamese MLP encoders and adding Gaussian noise to node embeddings instead of introducing complex operations over graphs. To verify the effectiveness of this new data augmentation fashion, we firstly conduct expensive ablation experiments in Fig. \ref{mlp_noise_ablation}. Here, we denote ``U'', ``G'', and ``U+G'', as the strategy of setting parameter un-shared MLPs, adding Gaussian noise to node embeddings, and both of them, respectively. From these results, we have two findings as follows. 1) These two simple strategies, which both aim to construct two different views of the same node, improve the clustering performance. 2) The combination of these two strategies achieve the best performance. In summary, we have verified the effectiveness of our proposed data augmentation fashion through these experimental results.

\begin{table}[!t]
\caption{Performance comparisons of different augmentations on six datasets. All results are reported with mean±std under ten runs. The bold values indicate the best results.}
\small
\centering
\scalebox{0.78}{
\begin{tabular}{c|c|cccc}
\hline
\multirow{2}{*}{\textbf{Dataset}}  & \multirow{2}{*}{\textbf{Metric}} & \multirow{2}{*}{\textbf{Drop}} & \multirow{2}{*}{\textbf{Add}} & \multirow{2}{*}{\textbf{Diffusion}} & \multirow{2}{*}{\textbf{Ours}} \\
                                   &                                  &                                      &                                     &                                     &                                \\ \hline
\multirow{4}{*}{\textbf{CORA}} & ACC                              & 72.03±0.93                           & 72.02±0.91                          &  72.94±0.75                    & \textbf{73.88±0.88}            \\
                                   & NMI                              & 54.85±0.95                           & 54.73±1.12                          & 55.83±1.14                    & \textbf{56.10±0.72}            \\
                                   & ARI                              & 49.71±1.65                           & 49.66±1.70                          &  51.44±1.40                    & \textbf{51.79±1.59}            \\
                                   & F1                               &  70.07±0.80                    & 69.86±0.96                          & 69.67±1.71                          & \textbf{70.81±1.96}            \\ \hline
\multirow{4}{*}{\textbf{CITESEER}}     & ACC                              & 67.74±0.51                           &  68.31±0.56                    & 68.15±0.49                          & \textbf{71.02±0.77}            \\
                                   & NMI                              & 42.86±0.59                           & 42.94±0.84                          &  43.12±0.53                    & \textbf{45.25±0.45}            \\
                                   & ARI                              & 43.28±0.65                           &  43.69±0.82                    & 43.40±0.92                          & \textbf{46.29±1.13}            \\
                                   & F1                               & 63.38±0.3                            &  63.54±1.17                    & 63.02±1.57                          & \textbf{64.80±1.01}            \\ \hline
\multirow{4}{*}{\textbf{AMAP}}     & ACC                              & 76.94±0.36                           & 76.78±0.37                          &  77.15±0.56                    & \textbf{77.48±0.37}            \\
                                   & NMI                              & 66.82±0.90                           & 66.41±1.30                          &  67.43±0.85                    & \textbf{67.67±0.88}            \\
                                   & ARI                              & 58.07±0.91                           & 57.81±0.62                          &  58.42±0.98                    & \textbf{58.48±0.72}            \\
                                   & F1                               & 71.65±0.47                           &  72.14±1.05                    & 71.23±0.75                          & \textbf{72.22±0.97}            \\ \hline
\multirow{4}{*}{\textbf{BAT}}      & ACC                              & 57.25±2.08                           & 67.63±1.04                          &  69.01±2.47                    & \textbf{77.94±0.99}            \\
                                   & NMI                              & 29.23±3.98                           & 42.51±1.20                          &  44.89±2.23                    & \textbf{52.91±0.68}            \\
                                   & ARI                              & 24.28±2.32                           & 37.47±1.93                          &  39.62±3.27                    & \textbf{50.64±1.85}            \\
                                   & F1                               & 56.46±2.86                           & 67.30±0.76                          &  68.36±2.81                    & \textbf{78.03±0.96}            \\ \hline
\multirow{4}{*}{\textbf{EAT}}      & ACC                              & 51.03±2.03                           &  57.37±0.26                    & 54.74±1.43                          & \textbf{57.94±0.42}            \\
                                   & NMI                              & 32.27±0.71                           &  32.62±0.39                   & 29.25±1.50                          & \textbf{33.91±0.49}            \\
                                   & ARI                              & 24.03±1.07                           &  26.27±0.74                   & 24.42±1.58                          & \textbf{27.51±0.59}            \\
                                   & F1                               & 47.99±1.74                           &  57.43±0.6                    & 53.98±1.82                          & \textbf{57.96±0.46}            \\ \hline
\multirow{4}{*}{\textbf{UAT}}      & ACC                              & 51.54±1.00                           & 53.61±0.11                          &  55.13±1.20                    & \textbf{56.58±1.62}            \\
                                   & NMI                              & 18.81±2.13                           & 24.59±1.07                          &  25.52±1.80                    & \textbf{28.07±0.71}            \\
                                   & ARI                              & 16.69±2.57                           & 18.10±0.86                          &  22.68±2.06                    & \textbf{24.80±1.85}            \\
                                   & F1                               & 49.89±1.31                           & 52.16±1.04                          &  54.08±1.31                    & \textbf{55.52±0.87}            \\ \hline
\end{tabular}}
\label{augment_ablation}
\end{table}

In addition, we compare our data augmentation fashion with other classical graph data augmentations including edge dropping \cite{SCAGC}, edge adding \cite{SCAGC}, and graph diffusion \cite{MVGRL,DCRN,IDCRN}. Concretely, in Table \ref{augment_ablation}, the data augmentation in SCGC is replaced by randomly dropping 10\% edges (``Drop''), or randomly adding 10\% edges (``Add'') or, graph diffusion (``Diffusion'') with 0.20 teleportation rate. From the results, we have two observations as follows. 1) The clustering performance is harmed by random edge dropping and adding, which may lead to semantic drift \cite{MoCL}. 2) Graph diffusion could achieve comparable performance on CORA and AMAP datasets while can not compare with ours on other datasets. It indicates that graph diffusion might change the underlying semantics of graphs \cite{AFGRL}. Overall, expensive experiments have demonstrated the effectiveness of our proposed data augmentation method.

\begin{figure}[!t]
\centering
\footnotesize
\begin{minipage}{0.48\linewidth}
\centerline{\includegraphics[width=1\textwidth]{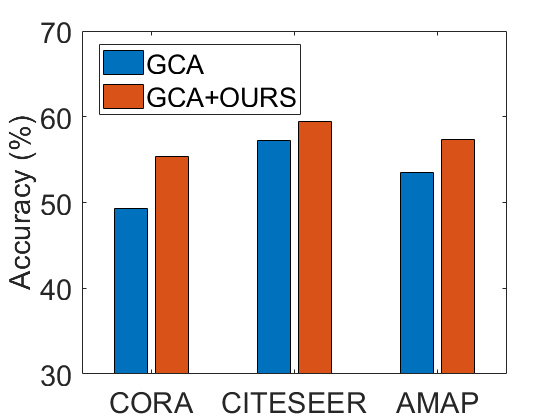}}
\vspace{1pt}
\centerline{\includegraphics[width=1\textwidth]{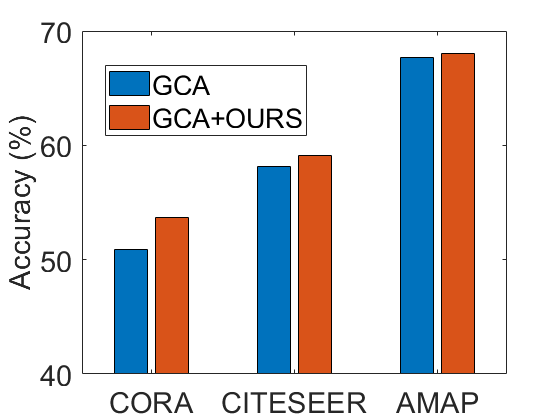}}
\vspace{1pt}
\centerline{\includegraphics[width=1\textwidth]{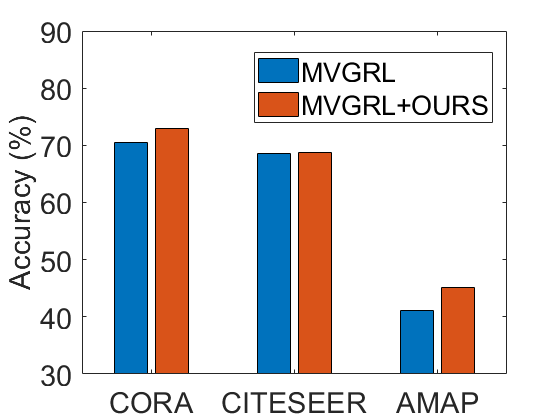}}
\vspace{1pt}
\end{minipage}
\begin{minipage}{0.48\linewidth}
\centerline{\includegraphics[width=1\textwidth]{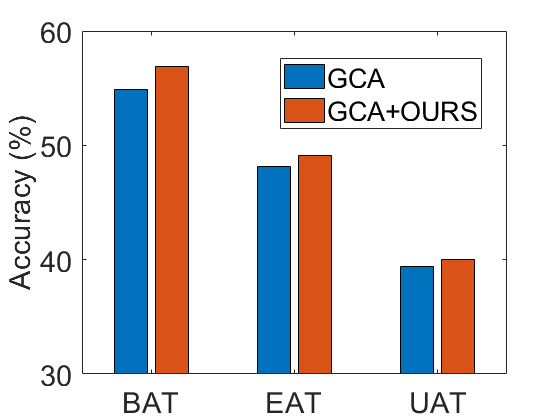}}
\vspace{1pt}
\centerline{\includegraphics[width=1\textwidth]{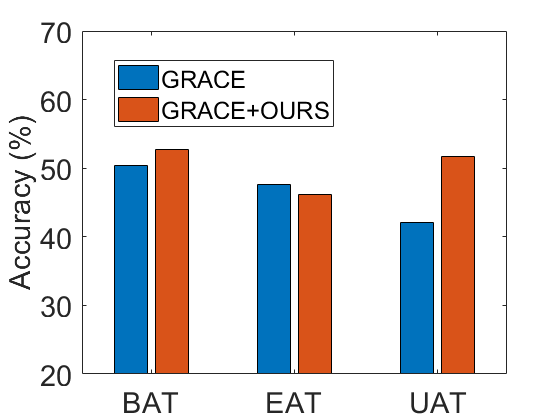}}
\vspace{1pt}
\centerline{\includegraphics[width=1\textwidth]{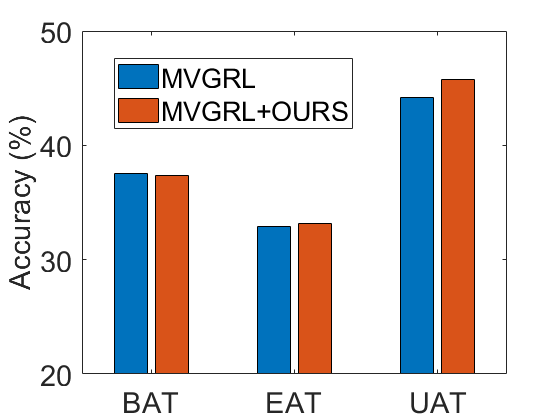}}
\vspace{1pt}
\end{minipage}
\caption{Experimental results of transferring our augmentation fashion to GCA \cite{GCA}, GRACE \cite{GRACE} and, MVGRL \cite{MVGRL}.}
\label{tranfer}
\end{figure}

To further verify the effectiveness and compatibility of our proposed data augmentation, we conduct experiments to transfer our augmentation fashion to other contrastive methods, including MVGRL \cite{MVGRL}, GRACE \cite{GRACE}, and GCA \cite{GCA}. Specifically, in these methods, we modify the original data augmentation fashion by setting un-shared siamese encoders and adding Gaussian noise to node embeddings. For the parameter settings and objective functions, we keep them consistent with the original literature. From the results in Fig \ref{tranfer}, we have two conclusions as follows. 1) Our proposed augmentation fashion is compatible with the existing contrastive methods. 2) Except for the performance of GRACE on EAT dataset and MVGRL on BAT dataset, the clustering performance of these contrastive methods could be improved by our proposed augmentation.

\begin{figure}[h]
\centering
\small
\begin{minipage}{0.49\linewidth}
\centerline{\includegraphics[width=1\textwidth]{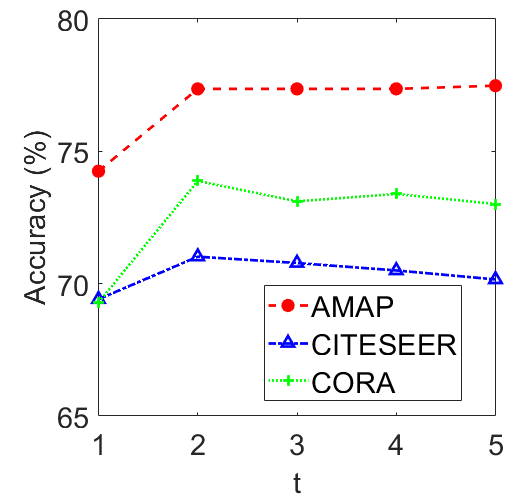}}
\end{minipage}
\begin{minipage}{0.49\linewidth}
\centerline{\includegraphics[width=1\textwidth]{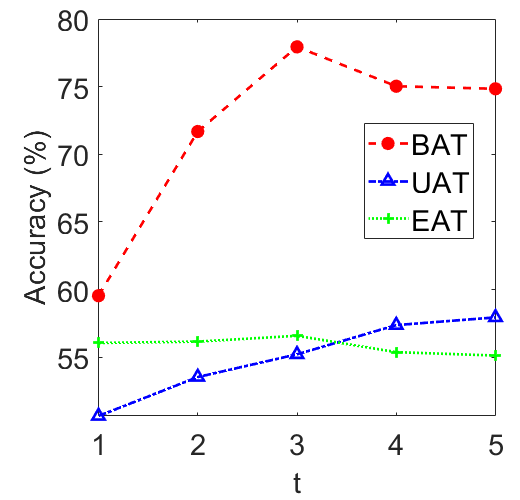}}
\end{minipage}
\caption{Sensitivity analysis of the layer number $t$ of graph Laplacian filters on six datasets.}
\label{Sensitivity_t}
\end{figure}

\subsection{Sensitivity Analysis of Hyper-parameters}
\subsubsection{\textbf{Sensitivity Analysis of hyper-parameter $t$}}
We conduct experiments to investigate the influence of the layer number $t$ of graph Laplacian filters on our proposed SCGC. As shown in Fig. \ref{Sensitivity_t}, we have two observations as follows. 1) SCGC could achieve promising performance when $t \in [2, 3]$. 2) Our proposed model becomes insensitive to $t$ when $3 \textless t \le 5$.


\begin{figure}[h]
\centering
\small
\begin{minipage}{0.49\linewidth}
\centerline{\includegraphics[width=1\textwidth]{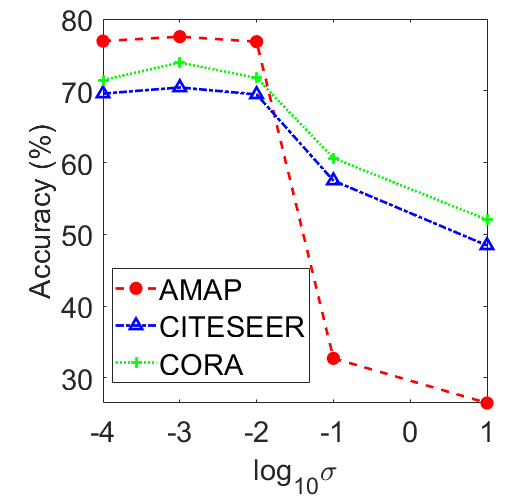}}
\end{minipage}
\begin{minipage}{0.49\linewidth}
\centerline{\includegraphics[width=1\textwidth]{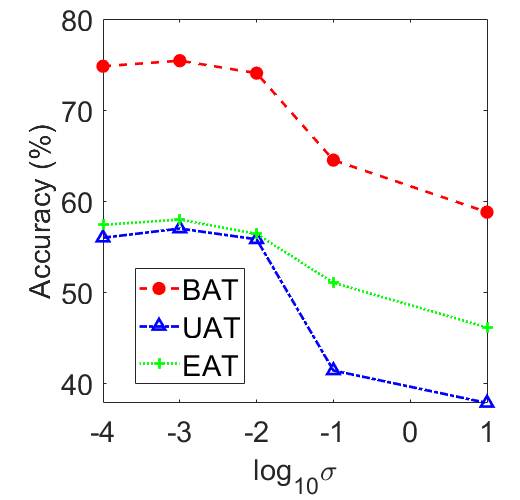}}
\end{minipage}
\caption{Sensitivity analysis of the standard deviation $\sigma$ of Gaussian noise on six datasets.}
\label{Sensitivity_sigma}`
\end{figure}

\begin{figure*}[!t]
\footnotesize
\begin{minipage}{0.139\linewidth}
\centerline{\includegraphics[width=\textwidth]{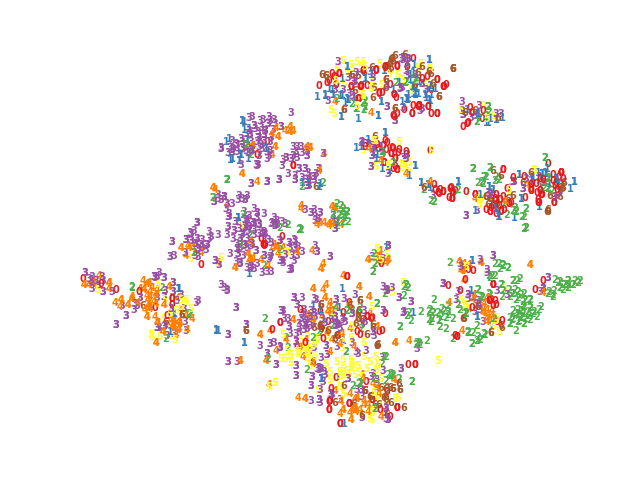}}
\vspace{3pt}
\centerline{\includegraphics[width=\textwidth]{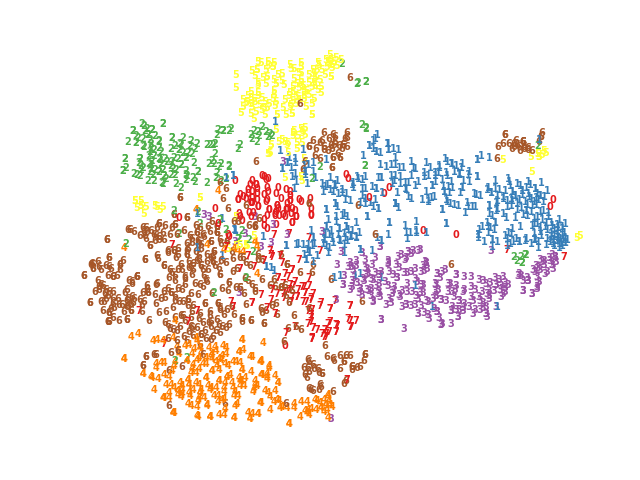}}
\vspace{3pt}
\centerline{AE}
\end{minipage}
\begin{minipage}{0.139\linewidth}
\centerline{\includegraphics[width=\textwidth]{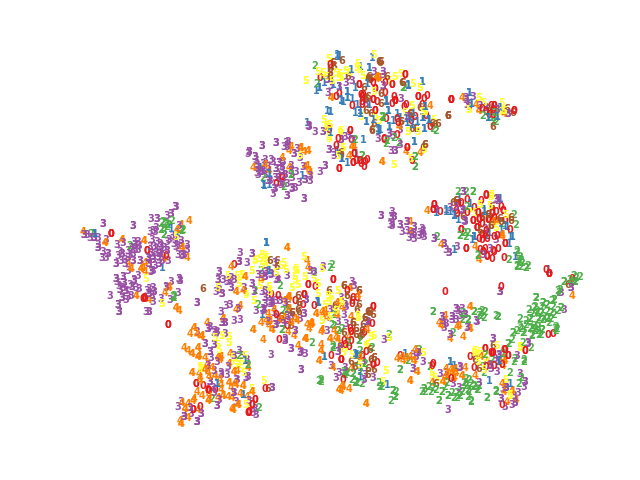}}
\vspace{3pt}
\centerline{\includegraphics[width=\textwidth]{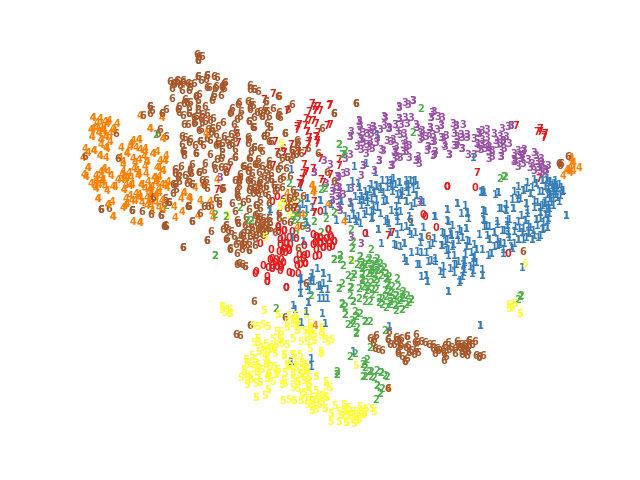}}
\vspace{3pt}
\centerline{DEC}
\end{minipage}
\begin{minipage}{0.139\linewidth}
\centerline{\includegraphics[width=\textwidth]{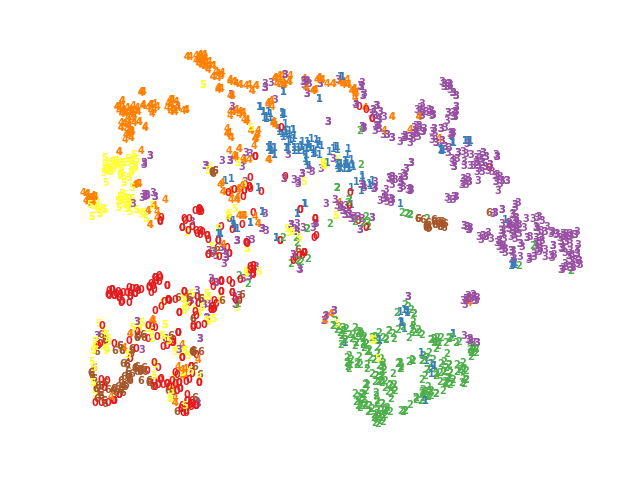}}
\vspace{3pt}
\centerline{\includegraphics[width=\textwidth]{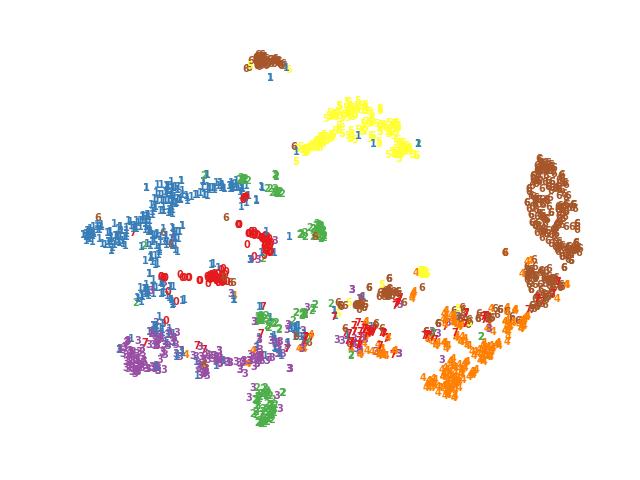}}
\vspace{3pt}
\centerline{GAE}
\end{minipage}
\begin{minipage}{0.139\linewidth}
\centerline{\includegraphics[width=\textwidth]{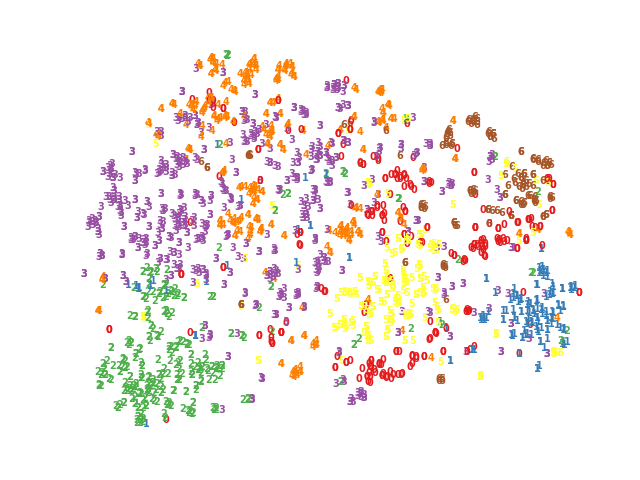}}
\vspace{3pt}
\centerline{\includegraphics[width=\textwidth]{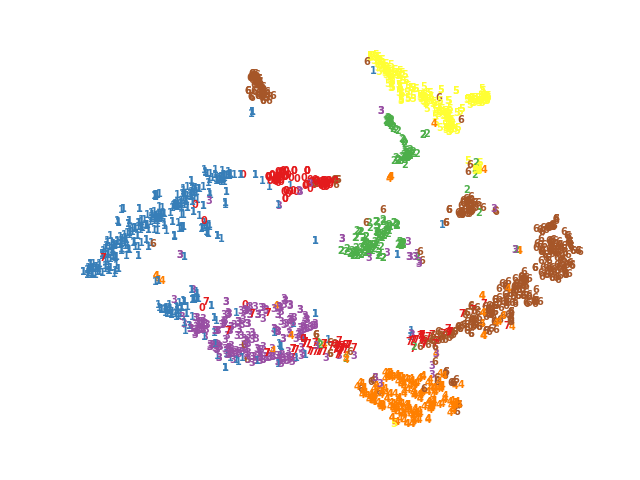}}
\vspace{3pt}
\centerline{DAEGC}
\end{minipage}
\begin{minipage}{0.139\linewidth}
\centerline{\includegraphics[width=\textwidth]{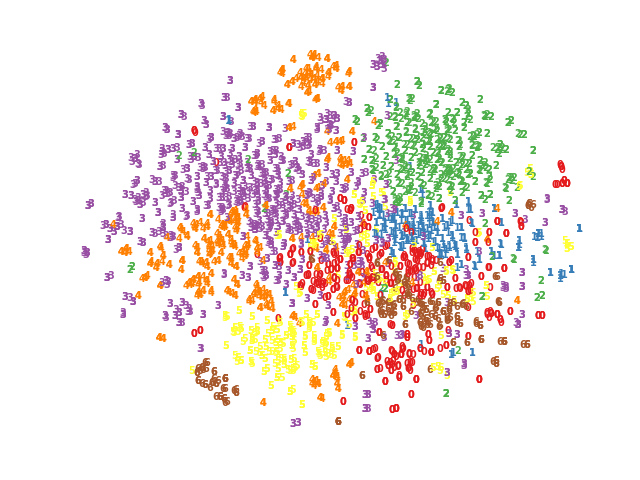}}
\vspace{3pt}
\centerline{\includegraphics[width=\textwidth]{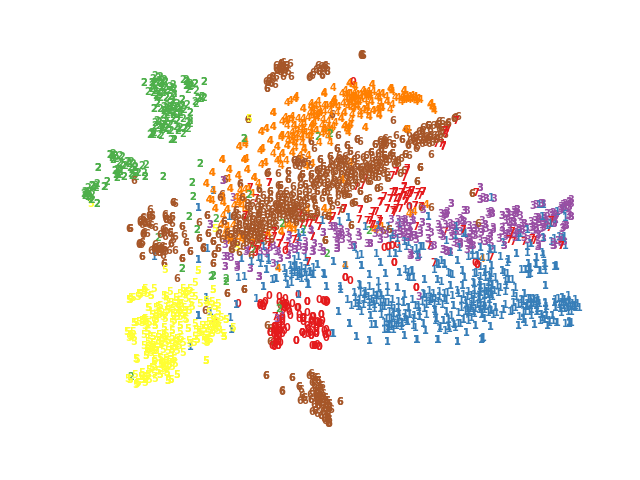}}
\vspace{3pt}
\centerline{MVGRL}
\end{minipage}
\begin{minipage}{0.139\linewidth}
\centerline{\includegraphics[width=\textwidth]{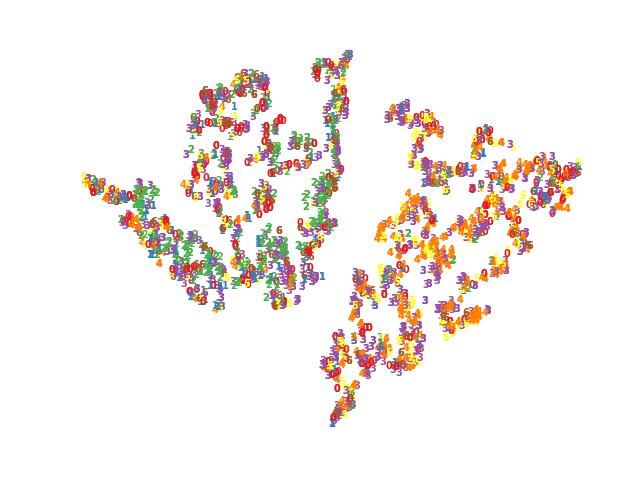}}
\vspace{3pt}
\centerline{\includegraphics[width=\textwidth]{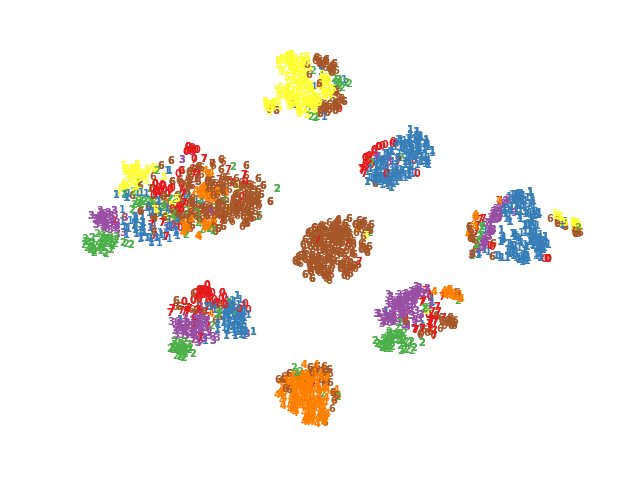}}
\vspace{3pt}
\centerline{SDCN}
\end{minipage}
\begin{minipage}{0.139\linewidth}
\centerline{\includegraphics[width=\textwidth]{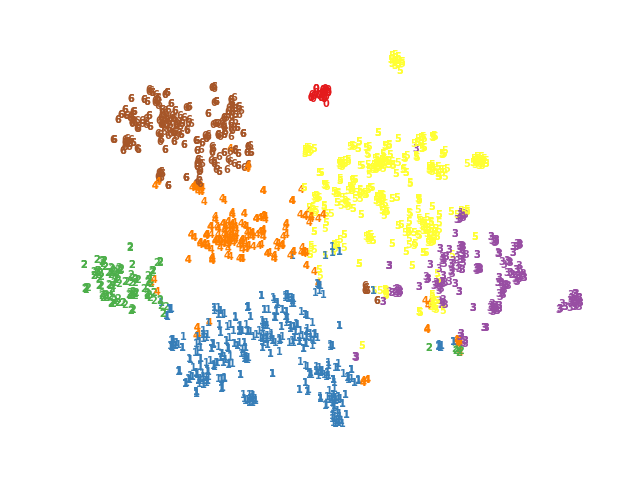}}
\vspace{3pt}
\centerline{\includegraphics[width=\textwidth]{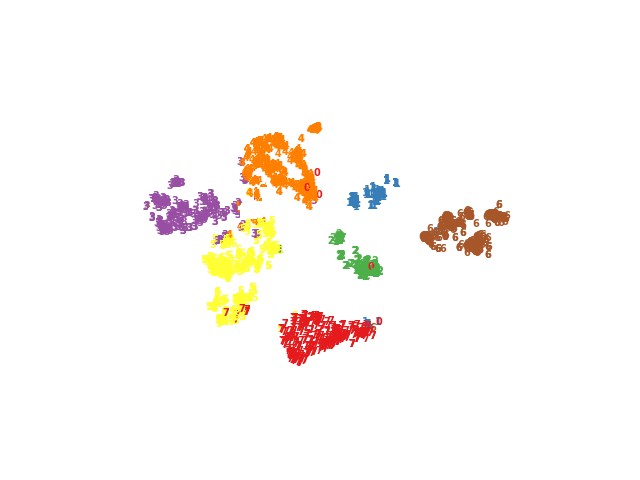}}
\vspace{3pt}
\centerline{SCGC (Ours)}
\end{minipage}

\caption{2D visualization on two datasets. The first row and second row correspond to CORA and AMAP, respectively.}
\label{t_SNE}  
\end{figure*}

\subsubsection{\textbf{Sensitivity Analysis of hyper-parameter $\sigma$}}
Besides, we investigate the robustness of our proposed method SCGC to hyper-parameter $\sigma$, which controls the Gaussian noise to the node embeddings $\textbf{Z}^{v_2}$. From the results in Fig. \ref{Sensitivity_sigma}, two conclusions are obtained as follows. 1) Our SCGC is robust to $\sigma$ when $\sigma \in [0.001, 0.1]$. 2) The clustering performance decreases drastically when $\sigma \textgreater 0.1$. The reason is that too much noise would lead to node embedding semantic drift. We set $\sigma$ to 0.01 in our model.



\subsubsection{Sensitivity Analysis of Layer Number of MLPs}
\begin{figure}[h]
\centering
\begin{minipage}{0.45\linewidth}
\centerline{\includegraphics[width=1\textwidth]{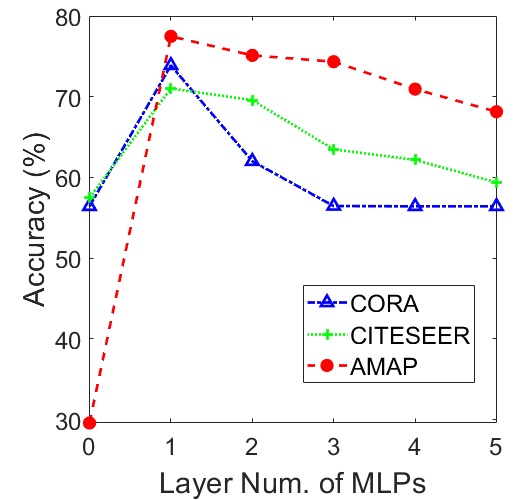}}
\end{minipage}
\begin{minipage}{0.45\linewidth}
\centerline{\includegraphics[width=1\textwidth]{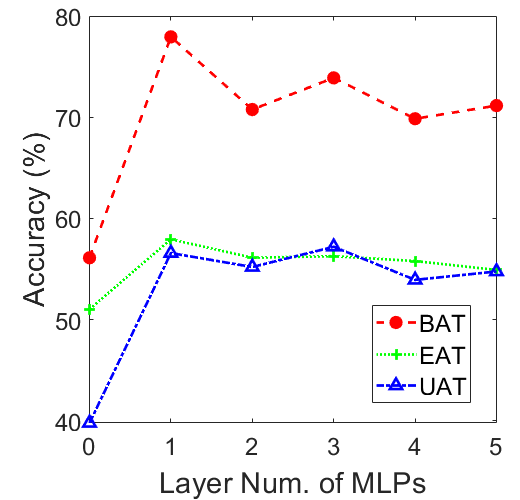}}
\end{minipage}
\caption{Analysis on the layer number of MLPs in our SCGC.}
\label{mlp_layer}
\end{figure}

We analyze the layer number of MLPs in our proposed method in this section. In Fig. \ref{mlp_layer}, it is worth mentioning that we directly perform the clustering algorithm on the smoothed attributes when the layer number of MLPs is equal to zero. From these results, we conclude as follows. 1) The MLP encoders are effective in improving clustering performance. 2) Our method could achieve the best performance when the layer number of MLPs equals one.

\subsection{Visualization Analysis}

To show the superiority of SCGC intuitively, we visualize the distribution of learned embeddings of SCGC and six compared baselines on CORA and AMAP datasets via $t$-SNE algorithm \cite{T_SNE}. As shown in Fig. \ref{t_SNE}, visible results demonstrate that SCGC better reveals the intrinsic clustering structure compared with other baselines. 

\section{CONCLUSION}
In this paper, we propose a contrastive deep graph clustering method termed Simple Contrastive Graph Clustering (SCGC) to improve the existing methods from the perspectives of network architecture, data augmentation, and objective function. As to the architecture, our network mainly includes two parts, i.e., pre-processing and network backbone. Concretely, a simple low-pass denoising operation conducts neighbor information aggregation as an independent pre-processing. Through this operation, we filter out the high-frequency noise in attributes in an efficient manner, thus improving the clustering performance. Besides, only two MLPs are included as the backbone. For data augmentation, we construct different graph views by setting parameter un-shared encoders and perturbing the node embeddings instead of introducing complex operations over graphs. Furthermore, we propose a novel neighbor-oriented contrastive loss to keep cross-view structural consistency, thus enhancing the discriminative capability of the network. Benefiting from the simplicity of SCGC, it is free from pre-training and saves both time and space for network training. Significantly, our algorithm outperforms the recent contrastive deep clustering competitors with at least seven times speedup on average. Extensive experimental results on seven datasets have demonstrated the effectiveness and superiority of SCGC. In the future, it is worth trying to design deep graph clustering methods for large-scale graph data.

\ifCLASSOPTIONcaptionsoff
  \newpage
\fi

\bibliographystyle{IEEEtran}
\bibliography{ref}

\end{document}